\documentclass[sigconf]{acmart}

\usepackage{comment}                
\usepackage{graphicx}               
\usepackage{subfigure}              
\usepackage{amsmath}                
\usepackage{algorithm}              
\usepackage{algorithmic}            
\usepackage{multirow}               
\usepackage{booktabs}               
\usepackage{setspace}               

\AtBeginDocument{%
  }

\copyrightyear{2026}
\acmYear{2026}
\setcopyright{cc}
\setcctype{by}
\acmConference[WSDM '26]{Proceedings of the Nineteenth ACM International Conference on Web Search and Data Mining}{February 22--26, 2026}{Boise, ID, USA}
\acmBooktitle{Proceedings of the Nineteenth ACM International Conference on Web Search and Data Mining (WSDM '26), February 22--26, 2026, Boise, ID, USA}
\acmDOI{10.1145/3773966.3777989}
\acmISBN{979-8-4007-2292-9/2026/02}

\begin{document}

\title{Sharpness-aware Federated Graph Learning}

\author{Ruiyu Li}
\authornote{Equal contribution.}
\email{ruiyli@stu.xidian.edu.cn}
\orcid{0009-0004-7813-7020}
\affiliation{%
  \institution{Xidian University}
  \city{Xi'an}
  \country{China}
}

\author{Peige Zhao}
\email{peigezhao@gmail.com}
\authornotemark[1]
\orcid{0000-0002-9346-9574}
\affiliation{%
  \institution{iFLYTEK}
  \city{Hefei}
  \country{China}
}

\author{Guangxia Li}
\authornote{Corresponding author.}
\email{gxli@xidian.edu.cn}
\orcid{0000-0003-2158-8723}
\affiliation{%
  \institution{Xidian University}
  \city{Xi'an}
  \country{China}
}

\author{Pengcheng Wu}
\email{pengcheng.wu@ntu.edu.sg}
\orcid{000-0003-0487-2060}
\affiliation{%
  \institution{Nanyang Technological University}
  \country{Singapore}
}

\author{Xingyu Gao}
\email{gxy9910@gmail.com}
\orcid{0000-0002-4660-8092}
\affiliation{%
  \institution{University of Chinese Academy of Sciences}
  \city{Beijing}
  \country{China}
}

\author{Zhiqiang Xu}
\email{zhiqiangxu2001@gmail.com}
\orcid{0000-0002-5693-8933}
\affiliation{%
  \institution{MBZUAI}
  \country{United Arab Emirates}
}

\renewcommand{\shortauthors}{Ruiyu Li et al.}

\begin{abstract}
One of many impediments to applying graph neural networks (GNNs) in processing large-volume real-world graph-structured data is that it disapproves of a centralized training scheme which involves gathering data belonging to different organizations due to privacy concerns. 
As a distributed data processing scheme, federated graph learning (FGL) enables learning GNN models collaboratively without sharing participants' private data. 
Though theoretically feasible, a core challenge in FGL systems is the variation of local training data distributions among clients, also known as the data heterogeneity problem. 
Most existing solutions suffer from two problems: 
(1) The typical optimizer based on empirical risk minimization tends to cause local models to fall into sharp valleys and weakens their generalization to out-of-distribution graph data. 
(2) The prevalent dimensional collapse in the learned representations of local graph data has an adverse impact on the classification capacity of the GNN model. 
To this end, we formulate a novel optimization objective that is aware of the sharpness (i.e., the curvature of the loss surface) of local GNN models. 
By minimizing the loss function and its sharpness simultaneously, we seek out model parameters in a flat region with uniformly low loss values, thus improving the generalization over heterogeneous data. 
By introducing a regularizer based on the correlation matrix of local representations, we relax the correlations of representations generated by individual local graph samples, so as to alleviate the dimensional collapse of the learned model. 
The proposed \textbf{S}harpness-aware f\textbf{E}derated gr\textbf{A}ph \textbf{L}earning (SEAL) algorithm can enhance the classification accuracy and generalization ability of local GNN models in federated graph learning. 
Experimental studies on several graph classification benchmarks show that SEAL consistently outperforms SOTA FGL baselines and provides gains for more participants. 
\end{abstract}




\maketitle

\section{Introduction}
\label{sec1}

Graph learning is a powerful tool to process graph-structured data. 
Most graph learning methods, especially graph neural networks (GNNs), are trained on large-scale centralized graph data. 
These methods have been extensively applied in various domains, including drug discovery~\cite{RBX20+, YWZ24+}, social networks~\cite{YZW20+, ZGP22+}, knowledge graphs~\cite{CWZ23, BCX23+}, and recommender systems~\cite{YPZ20+, MRX22+}. 
However, in many real-world scenarios, graph data are geographically distributed and cannot be directly shared due to privacy concerns. 
For example, in bioinformatics, graph classification techniques are employed to learn protein representations and classify them as enzymes or non-enzymes. 
These private protein molecules typically reside with multiple individual owners (i.e., clients). 
Due to intellectual property concerns and the intrinsic commercial value of molecules, academic labs, national labs, and private organizations are often unable to share their molecular data, leading to the data isolation problem. 

\begin{figure}[htbp]
    \centering
    \subfigure[Intra-domain heterogeneity]{
    \includegraphics[width=0.45\linewidth]{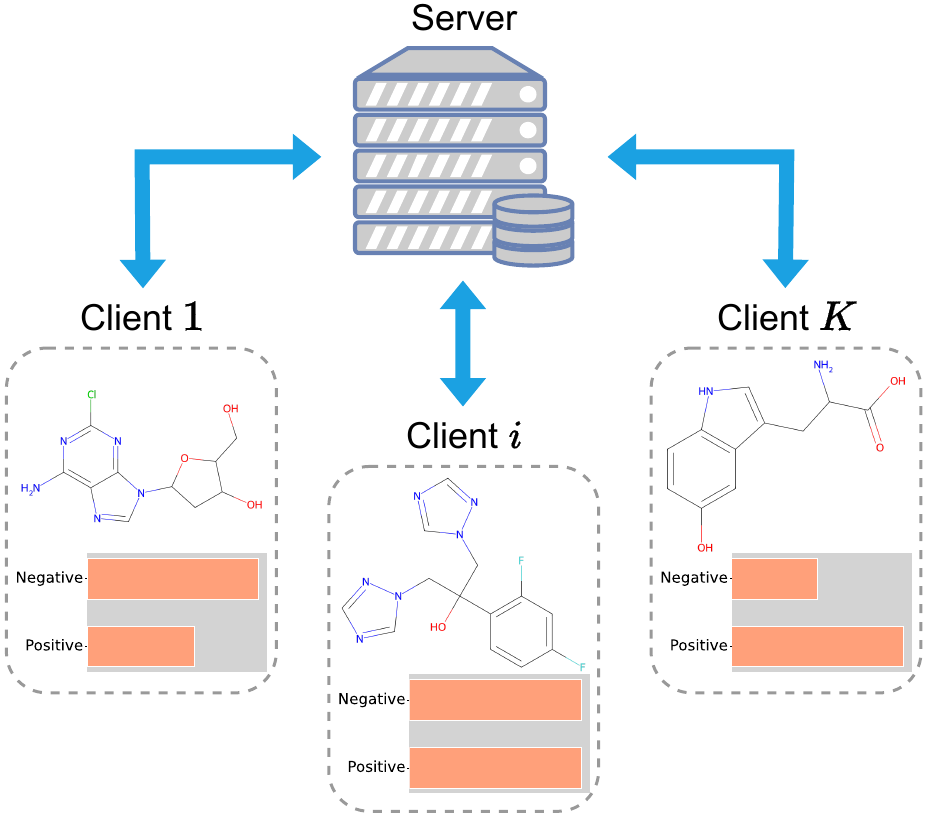}
    \label{fig:fig1a}
    }
    \subfigure[Inter-domain heterogeneity]{
    \includegraphics[width=0.45\linewidth]{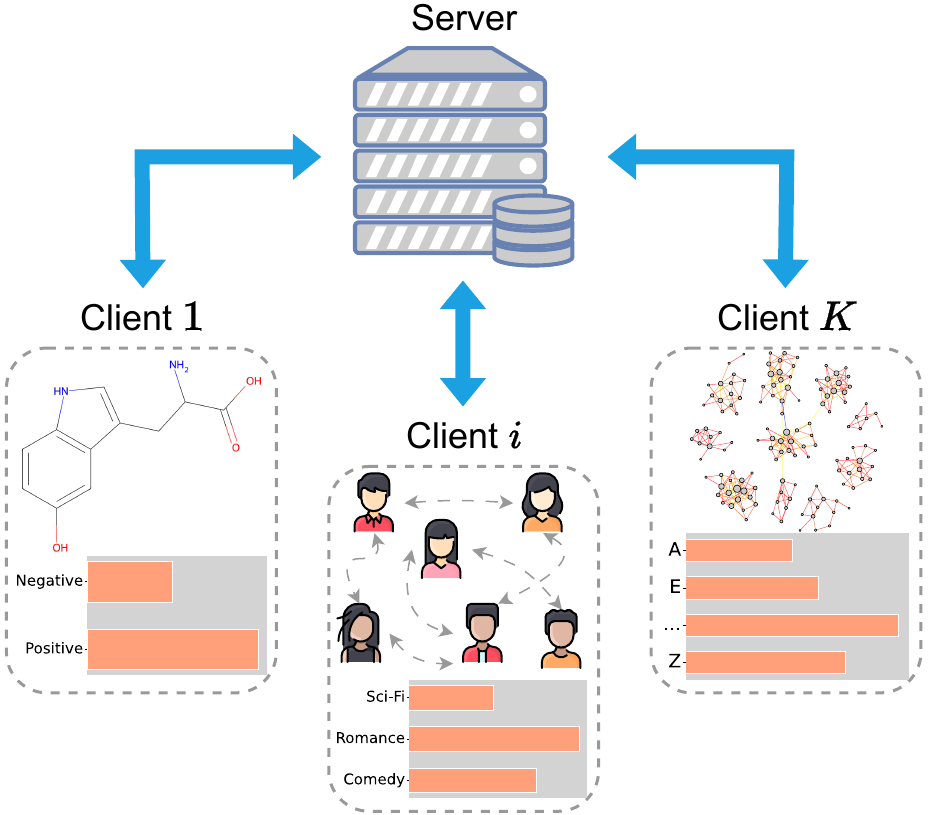}
    \label{fig:fig1b}
    }
    \caption{Graph data heterogeneity. ~\subref{fig:fig1a} Intra-domain heterogeneity involves two cases: first, clients possess data drawn from the same dataset but with inconsistent statistical properties; second, clients' data are sourced from different datasets within the same domain. ~\subref{fig:fig1b} Inter-domain heterogeneity implies that clients' data are derived from entirely different domains.}
    \label{fig:fig1}
\end{figure}

Federated graph learning (FGL), a distributed learning paradigm for collaboratively training GNN models, has emerged as a promising direction for exploiting GNNs on decentralized graph data~\cite{TLL23+, HWY23+}. 
A major challenge in FGL is the underlying discrepancy in graph data distribution across clients, commonly referred to as graph data heterogeneity. 
As illustrated in Fig.~\ref{fig:fig1}, the heterogeneity manifests both intra and inter domains. 
Intra-domain heterogeneity arises when clients hold graph data from the same dataset or domain (e.g., small molecules) but with distinct statistical properties. 
In contrast, inter-domain heterogeneity occurs when each client's graph data is drawn from different domains, leading to discrepancies in both statistical characteristics and graph structural information. 
Such heterogeneity can severely destabilize the training process and degrade performance. 
While traditional federated learning approaches can effectively alleviate statistical inconsistency in computer vision tasks~\cite{MMR17+, KKM20+, LSZ20+, WLL20+}, they fall short when it comes to handling heterogeneous graph data. 

\begin{figure*}[htbp]
    \centering
    \subfigure[Non-IID, w/o RepDec]{
    \includegraphics[width=0.23\linewidth]{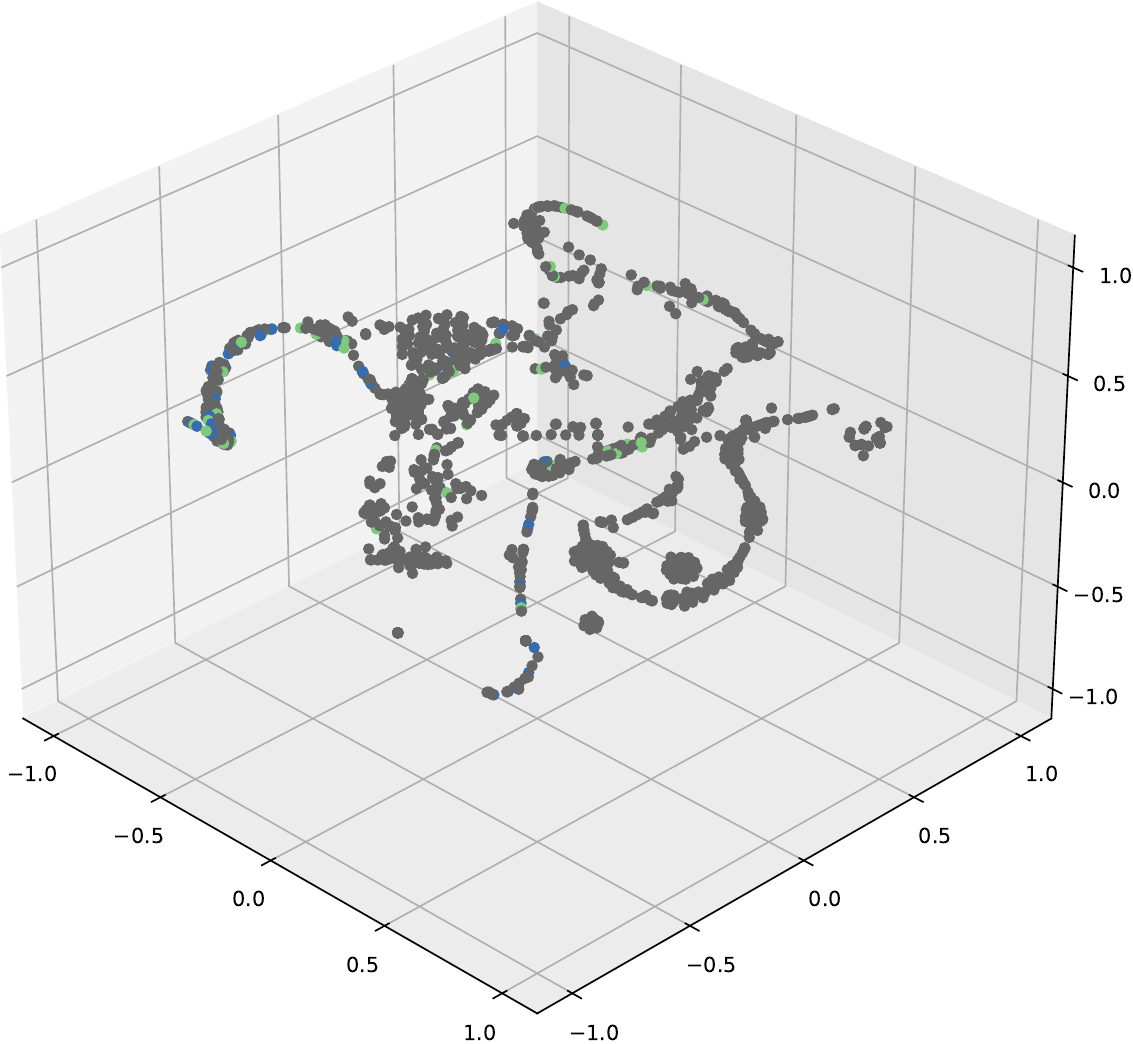}
    \label{fig:fig2a}
    }
    \subfigure[Non-IID, w/ RepDec]{
    \includegraphics[width=0.23\linewidth]{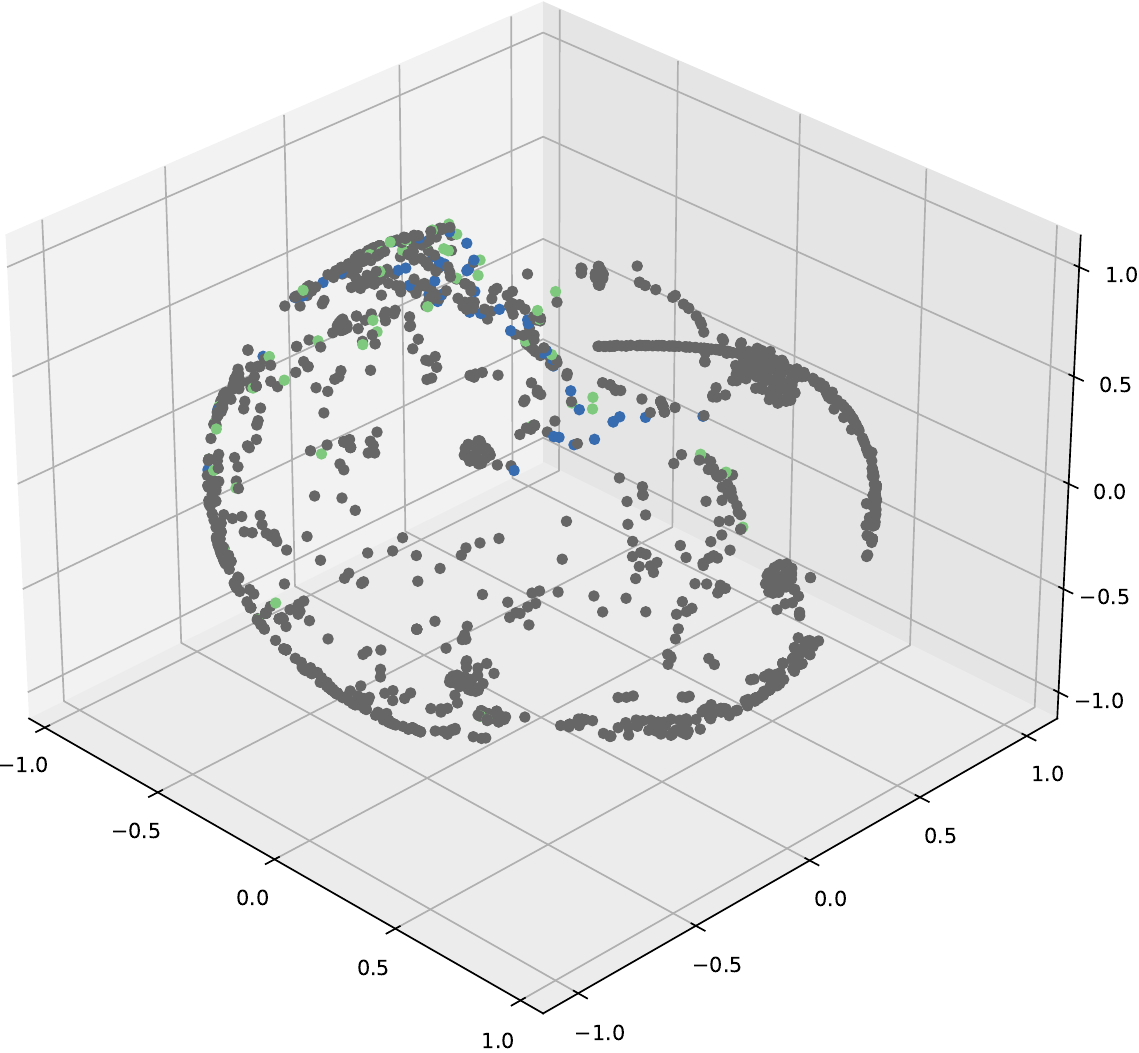}
    \label{fig:fig2b}
    }
    \subfigure[Inter-domain, w/o RepDec]{
    \includegraphics[width=0.23\linewidth]{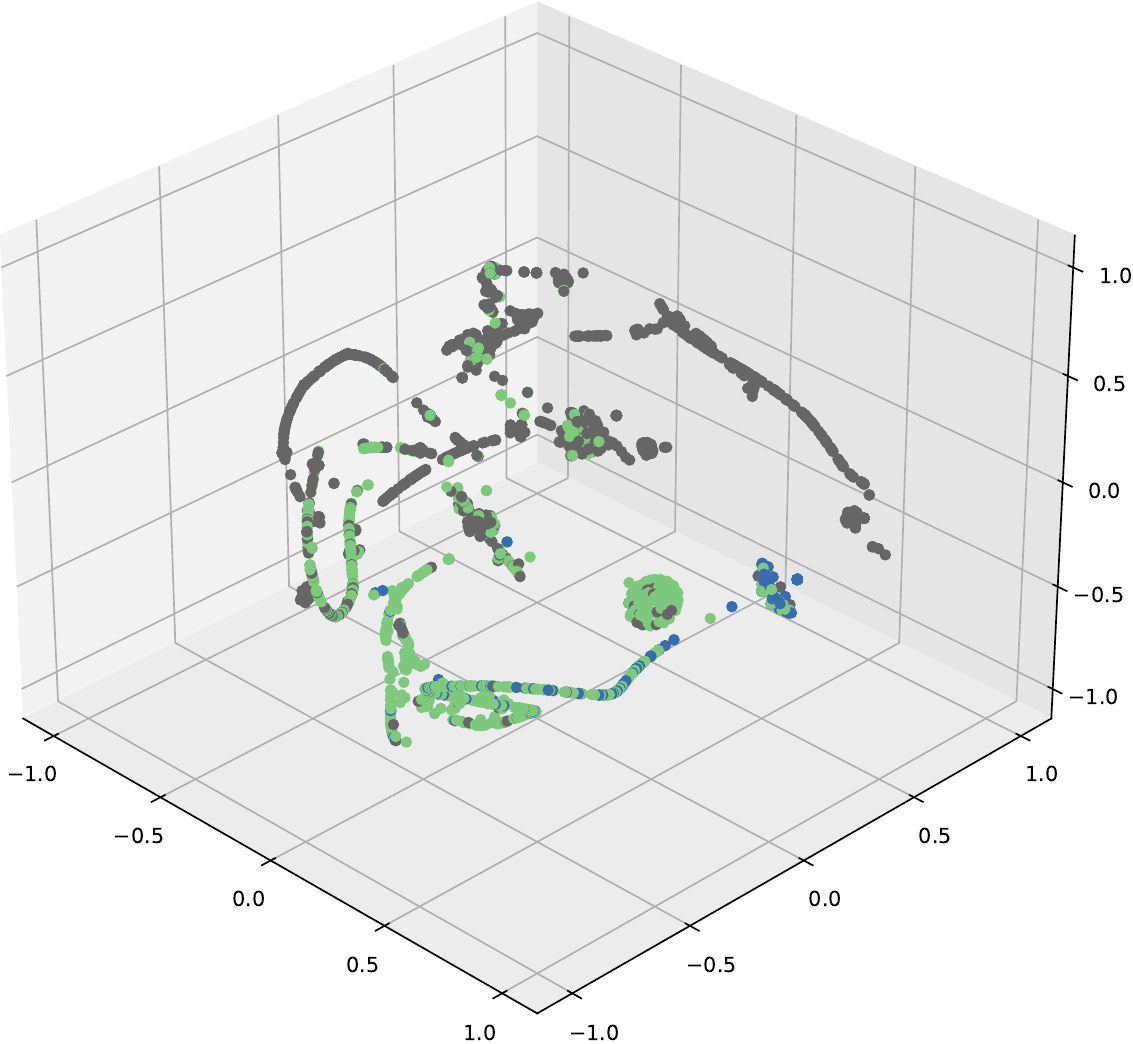}
    \label{fig:fig2c}
    }
    \subfigure[Inter-domain, w/ RepDec]{
    \includegraphics[width=0.23\linewidth]{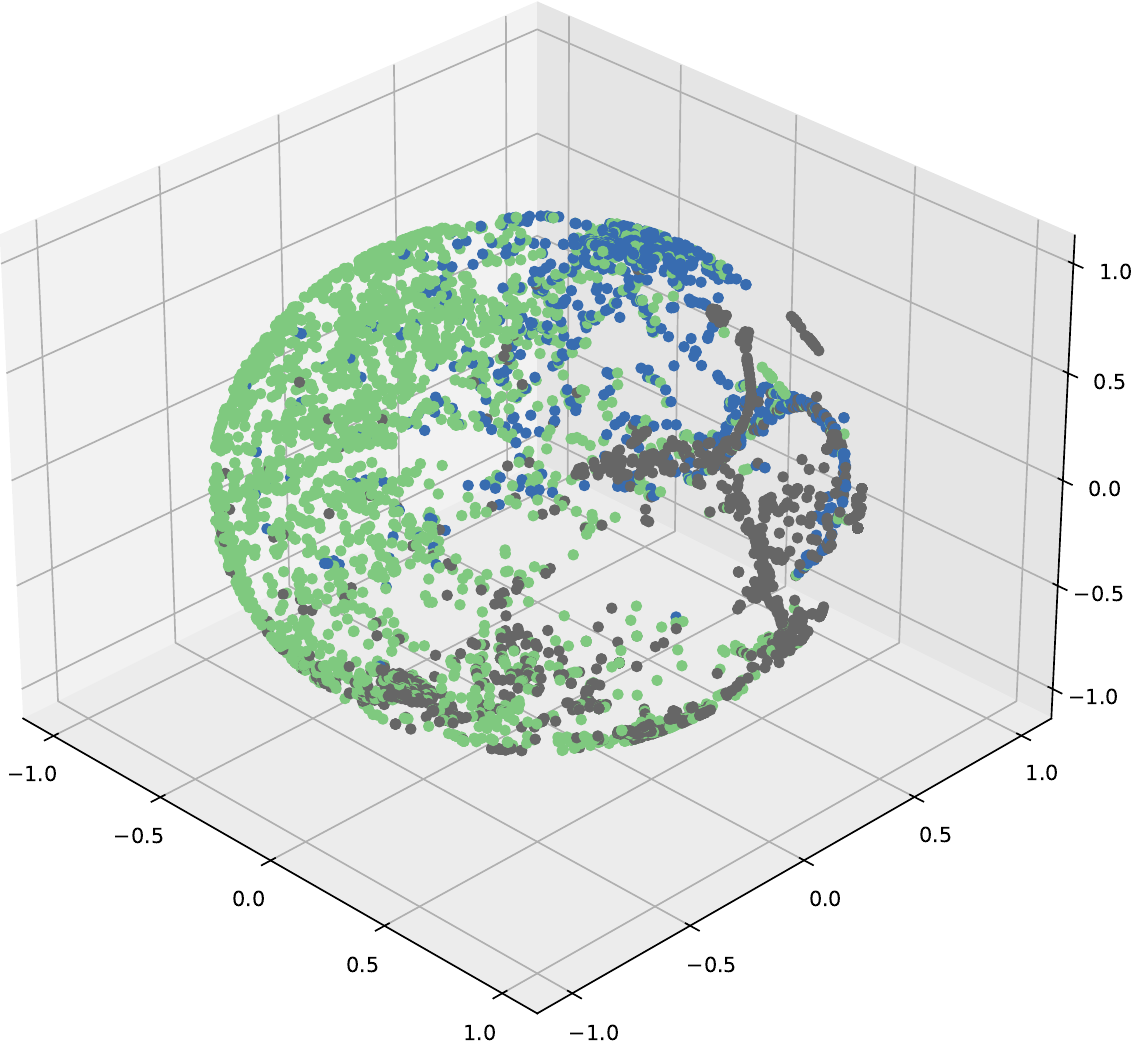}
    \label{fig:fig2d}
    }
    \caption{~\subref{fig:fig2a} and~\subref{fig:fig2b} show the representations generated by a randomly sampled client w/o and w/ representation decorrelation (RepDec) under the non-IID setting (COLLAB dataset), respectively. ~\subref{fig:fig2c} and~\subref{fig:fig2d} show the representations under the inter-domain setting (BioSnCV), respectively. The representations are all generated by a three-layer graph attention network and then normalized into a unit ball.}
    \label{fig:fig2}
\end{figure*}

Prior studies have sought to address the graph data heterogeneity problem from multiple perspectives, including optimization-based methods~\cite{LSZ20+, KKM20+}, personalized FGL~\cite{ZMC23+, TLL23+}, and clustered FGL~\cite{CMG21+, XMX21+}. 
For example, FedEgo~\cite{ZMC23+} and FedStar~\cite{TLL23+} share graph structural information across clients and introduce personalized feature knowledge learning modules on each client. 
FedCG~\cite{CMG21+} and GCFL+~\cite{XMX21+} group clients into non-overlapping clusters, within which clients share model parameters. 
However, personalized approaches introduce additional trainable parameters, while clustered approaches are highly sensitive to the quality of client clustering. 
Since GNNs are typically over-parameterized~\cite{ZSC23}, all of these methods incur substantial computational cost and communication overhead~\cite{LDC22+}, and they still suffer from the following issues: 
\begin{enumerate}
    \item GNN models trained under the empirical risk minimization (ERM) based optimizers tend to overfit local graph data and fail to generalize to out-of-distribution data. 
    \item Heterogeneous graph data can cause local GNN models to undergo dimensional collapse in the representation space (as illustrated in Fig.~\ref{fig:fig2a} and Fig.~\ref{fig:fig2c}). 
\end{enumerate}

To address the first issue, we note that models trained with ERM-based optimizers may fall into the sharp valley of the loss surface~\cite{FKM21+}, which leads to poor generalization, even when the training and test data distributions are similar. 
Previous efforts~\cite{DPB17+, PAS17+, NDJ17+} have explored the relationship between the flatness of local minima and the generalization ability of deep models to improve generalization and have applied it within federated learning~\cite{CCC22}, but their application to FGL has not been thoroughly investigated. 
In FGL, each local GNN model only has access to a subset of the global graph data and thus tends to overfit its local data to minimize the training loss, which harms its ability to generalize to the test data. 
This study investigates the higher-order information underlying GNN generalization, namely, the curvature of the loss surface, to enhance the generalization ability of as many local GNN models as possible. 
To this end, we equip each local GNN model with a sharpness-aware minimization optimizer that jointly optimizes for low training loss and a flatter loss landscape. 
The objective is to identify model parameters that reside in a neighborhood with uniformly low loss values, rather than at a single point that simply minimizes the loss, which naturally leads to a min--max optimization problem. 
We solve it via a two-step procedure: (1) applying gradient ascent to find a perturbation point that increases the loss within a fixed-radius neighborhood of the current parameters, and (2) updating the current parameters by gradient descent using the gradient evaluated at this perturbed point. 

To address the second issue, we observe a phenomenon of dimensional collapse in the representation space of local GNNs, particularly under heterogeneous graph data distributions (Fig.~\ref{fig:fig2}).
Specifically, graph representations learned by a three-layer graph attention network (GAT)~\cite{VCC18+} exhibit severe collapse, forcing them to reside in a low-dimensional space and impeding class discriminability.
In the FGL framework, the periodic aggregation and subsequent broadcasting of GNN backbones exacerbate this degradation\footnote{In this study, we aggregate only the GNN backbones rather than entire models, as inter-domain graph data involve distinct label spaces, necessitating task-specific classifiers tailored to local distributions.}.
To mitigate this effect, we propose to decorrelate local graph representations.
Since the singular values of the representation covariance matrix offer a rigorous characterization of their high-dimensional distribution, we introduce a regularization term to the local optimization objective to suppress these correlations.
Considering the high computational complexity of full singular value decomposition, we instead normalize the covariance matrix into a correlation matrix and minimize its norm.
This approach effectively encourages the model to learn more diverse and discriminative representations. 

In this study, we propose SEAL, a \textbf{S}harpness-aware f\textbf{E}derated gr\textbf{A}ph \textbf{L}earning algorithm. 
It integrates a sharpness-aware minimization (SAM) optimizer with a covariance-based decorrelation regularizer into the local training objective. 
Extensive evaluations across diverse graph classification benchmarks spanning four different domains demonstrate that SEAL consistently outperforms state-of-the-art FGL baselines, delivers excellent generalization, and adapts effectively to heterogeneous graph data distributions. 

\section{Related Work}
\label{sec2}

\subsection{Data Heterogeneity in Federated Learning}
Heterogeneous data can severely hinder both the convergence and generalization of federated models. 
Existing efforts attempt to alleviate the impact of data heterogeneity from three perspectives: local optimization, data resampling, and model aggregation. 
From the local optimization perspective, several studies optimize local models through well-designed regularization objectives~\cite{LSZ20+, AZM21+, LHS21, GLF22} and local model bias corrections~\cite{KKM20+, WLL20+, MS21, DBJ22}. 
For example, FedProx~\cite{LSZ20+} augments the local training objective with a proximal term. 
Based on FedProx, FedADMM~\cite{GLF22} introduces dual variables on the client side and incorporates them into the local objective, then optimizes this objective based on the distributed alternating direction method of multipliers~\cite{LZL25+}. 
SCAFFOLD~\cite{KKM20+} reduces client drift by maintaining control variates on each client and using them to correct local updates. 
From the data perspective, class-balanced resampling~\cite{HQB20, WXW21+, CC22} improves when client data are label-imbalanced, while data augmentation~\cite{YSH21+} has been shown to alleviate data heterogeneity in federated learning. 
From the model aggregation perspective, comparing local gradients with the global gradient, filtering out harmful client updates, and prioritizing clients that contribute more can accelerate convergence and reduce the negative impact of data heterogeneity~\cite{WW22, FVK21+}. 

\subsection{Federated Graph Learning}
Federated graph learning (FGL) enables distributed training of graph neural networks. 
Existing FGL methods can be broadly categorized into two settings: intra-graph and inter-graph. 
In the intra-graph setting, each client holds a subgraph of a larger global graph and typically tackles tasks such as node/link prediction~\cite{ZYL21+} and community detection~\cite{BJJ23+}. 
In the inter-graph setting, each client owns a collection of private graphs and collaboratively learns a GNN with improved generalization~\cite{XMX21+}. 
However, both settings suffer from the data heterogeneity. 
Directly applying classical federated learning methods to FGL often leads to significant performance degradation~\cite{HBC21+}. 
To address this, many studies either reduce the discrepancy between global and local GNN models via carefully designed optimization objectives~\cite{HWL22+, ZLW22} or share structural graph information across clients~\cite{HWY23+, TLL23+}.
For example, FedGCN~\cite{HWL22+} utilizes Top-K graph pooling layers and fully connected layers to extract node features, and introduces an online adjustable attention mechanism with a learnable parameter that assigns attention scores to local models, thereby mitigating the impact of low-quality local updates on the global model and improving fault tolerance and accuracy. 
FLIT+~\cite{ZLW22} reweights clients' training processes by assigning larger weights to uncertain graph data. 
FedStar~\cite{TLL23+} extracts and shares structural graph information across clients and then learns personalized feature-level knowledge on each client. 

\subsection{Sharpness-aware Minimization}
Sharpness-aware minimization (SAM)~\cite{FKM21+} optimizers have played a crucial role in improving deep neural network training by jointly minimizing the training loss and the sharpness of the loss landscape. 
Existing studies on SAM can be categorized into two lines: (1) enhancing the generalization ability of over-parameterized models and (2) improving the optimization efficiency of SAM. 
To improve generalization, a series of SAM-based methods have been proposed~\cite{KKP21+, ZGY22+, ZZH22, ZXY23+, YJY23+, BAS23+, LG23, LZH24+, BAD24+}, which refine either the sharpness measure or the associated optimization procedure. 
For example, GSAM~\cite{ZGY22+} enhances generalization by minimizing both the perturbed loss and a surrogate gap that measures the difference between the maximum loss in a neighborhood and the loss at the center point. 
GAM~\cite{ZXY23+} introduces a first-order flatness measure based on the maximum gradient norm within a prescribed perturbation radius, aiming to locate minima with uniformly small curvature. A key limitation of SAM is its computational cost: each optimization step requires two forward and two backward passes, which constrains its applicability at scale.
To address this, several efficient SAM variants have been proposed~\cite{LMC22+, DYF22+, NCG22+, ZZH22+, BAR24}.
For instance, LookSAM~\cite{LMC22+} reduces training cost by lowering the frequency of inner gradient computations, while maintaining accuracy. 
ESAM~\cite{DYF22+} accelerates SAM via two strategies: stochastic weight perturbation, which approximates sharpness by perturbing only a subset of weights, and sharpness-sensitive data selection, which optimizes the loss on carefully selected data samples. 
In this study, we focus on leveraging SAM to enhance the generalization ability of local GNN models, and leave the integration of accelerated SAM variants for GNN training to future work. 

\section{Preliminaries}
\label{sec3}

\subsection{Graph Neural Networks}
Let $G = (\mathcal{V}, \mathcal{E})$ be a graph composed of a set of nodes $\mathcal{V}$ and a set of edges $\mathcal{E}$ connecting these nodes. 
Each node $v \in \mathcal{V}$ is associated with a feature vector $x_v$. 
A GNN aims to learn representations that capture both the graph structure and node features, including node representation $h_v$ for each node $v \in \mathcal{V}$ and graph representation $h_G$ for the entire graph $G$. 
It typically consists of a message propagation and a neighbor aggregation module, where each node $v$ iteratively collects and aggregates representations from its neighbors with its own. 
Formally, for an $L$-layer GNN $f$, the update at layer $l$ is given by
\begin{equation}
    h_v^{l+1} = \sigma \left[W h_v^l+U\mathrm{Agg}\left( \left\{ h_u^l; u \in \mathcal{N}_v \right\} \right) \right] , \forall l \in \left[ L \right], \label{eq:eq1}
\end{equation}
where $h_v^{l}$ is the representation of node $v$ at the $l$-th layer, $h_v^{0} = x_v \in \mathbb{R}^{d_{v}}$ is the input node feature, $\mathcal{N}_v$ denotes the set of neighbors of $v$, $\mathrm{Agg}(\cdot)$ is an aggregation function (e.g., mean), $W, U\in \mathbb{R}^{d_{v}\times d_{v}}$ are learnable parameters and $\sigma$ represents an activation function (e.g., ReLU). 
To obtain the graph representation $h_G$, a $\mathrm{Readout}$ function is applied:  
\begin{equation}
    h_G = \mathrm{Readout} \left( \left\{ h_v^{L+1}; v \in \mathcal{V} \right\} \right), \label{eq:eq2}
\end{equation}
where $\mathrm{Readout}(\cdot)$ aggregates all node representations. 
Common choices for $\mathrm{Readout}(\cdot)$ include sum pooling, mean pooling, and max pooling. 

\subsection{Federated Learning}
In federated learning, $N$ clients are connected to a central server responsible for model aggregation. 
The $i$-th client ($i \in [N]$) owns a private dataset $\xi_i = (X_i, Y_i)$ consisting of $M_i$ data instances sampled from the distribution $\mathcal{D}_i$. 
Due to data heterogeneity, these distributions $\mathcal{D}_i$ differ across clients. 
Let $\mathcal{L}_i(w_i)$ denote the training loss on client $i$: $\mathcal{L}_i(w_i) := \sum_{\xi_{i, j}} \ell(w_i, \xi_{i, j}) $, where $\ell(w_i, \xi_{i, j})$ is the loss value of the $j$-th data instance and $w_i$ represents the parameters of the local model $f_i$ on client $i$. 
Federated learning performs the following global optimization by solving the ERM problem on each client: 
\begin{equation}
    \min_{\left(w_1, w_2, \cdots, w_N\right)} \frac{1}{N} \sum_{i \in [N]} \mathcal{L}_i (w_i). \label{eq:eq3}
\end{equation}

The goal of federated learning is to learn a globally shared model $f$ with parameter $w = w_1 = w_2 = \cdots = w_N$. 
The most well-known approach, FedAvg~\cite{MMR17+}, periodically aggregates local models on the server via 
\begin{equation}
    \bar{w} \leftarrow \sum_{i=1}^N \frac{M_i}{M} w_i, \label{eq:eq4}
\end{equation}
and then broadcasts the aggregated model $\bar{w}$ back to all clients, where $M = \sum_{i} M_i$ is the total number of data instances across all clients. 
Each client receives $\bar{w}$ from the server and performs local stochastic gradient descent (SGD) in parallel. 

In this study, we focus on inter-domain federated graph classification, where the number of graph classes differs across domains. 
As a consequence, client-specific classifiers cannot be directly shared. 
Instead, we only share the GNN backbone responsible for generating graph representations. 

\section{The SEAL Algorithm}
\label{sec4}

\subsection{Federated Graph Learning with Sharpness-aware Minimization}
Compared with general federated learning, federated graph learning is tailored to graph-structured data and faces several unique challenges, especially when handling graphs from different domains. 
The key challenges include: 
\begin{enumerate}
    \item Graph heterogeneity: Graph data across clients can differ substantially in both structural patterns and feature distributions, making GNN training and optimization more challenging. 
    \item Label imbalance: Graph data on different clients often exhibit highly imbalanced class distributions, which degrades local performance and weakens the generalization ability. 
    \item Inter-domain generalization: Clients process graph data from diverse domains and at different scales, requiring local GNN models to capture domain-invariant graph features that generalize well across clients. 
\end{enumerate}

In FGL, the graph data on each client are drawn from a local distribution $\mathcal{D}_i$, which is shifted from a mixture of an unknown distribution $\mathcal{D}$, i.e., $\mathcal{D}_i \neq \mathcal{D}$. 
Consequently, graph datasets exhibit distinct statistical properties across different clients $i$ and $j$ ($i \neq j$), implying $\mathcal{D}_i \neq \mathcal{D}_j$ and leading to graph data heterogeneity. 
Moreover, with an ERM-based optimizer, the $i$-th client typically converges to a local minimizer $w_i^*$ that fits its local distribution $\mathcal{D}_i$ to obtain the lowest loss, but this minimizer often lies in a region with a sharp loss surface~\cite{CCS17+}, leading to poor generalization of the local GNN model $f_i$. 
To improve the generalization of as many local GNN models as possible and better accommodate out-of-distribution graph data, we equip each client with an adaptive SAM optimizer. 

Specifically, the SAM optimizer aims to locate flat regions of the loss landscape with uniformly low loss values by introducing small perturbations to the parameters of local GNN models, thereby exploiting the connection between loss surface geometry and model generalization. 
It has been shown, both theoretically and empirically, to outperform standard ERM-based optimizers~\cite{FKM21+, KMN17+}. 
Leveraging the linear property of the FGL objective in Eq.~\eqref{eq:eq3}, we propose SEAL, a sharpness-aware federated graph learning algorithm that applies SAM to each client to minimize its perturbed loss. 
The optimization problem of SEAL is formulated as
\begin{equation}
    \min_{ (w_1, \cdots, w_N) } \max_{\parallel A_{w_i}^{-1} \epsilon_i \parallel_2^2 \leq \rho} \frac{1}{N} \sum_{i \in [N]} \left( \mathcal{L}_i (\tilde{w}_i) + \frac{\lambda}{2} \parallel w_i \parallel_2^2 \right), \label{eq:eq5}
\end{equation}
where $\tilde{w}_i = w_i + \epsilon_i$, $\epsilon_i$ is an adversarial perturbation of $w_i$, $\rho$ is the perturbation radius (a predefined hyper-parameter), $\lambda$ is the weight decay coefficient, and $\parallel \cdot \parallel_2$ denotes the $L 2$ norm. 

Note that $\left\{ A_{w_i}, w_i \in \mathbb{R}^k \right\}$ denotes a family of invertible linear operators on $\mathbb{R}^k$. 
The operator $A_{w_i}^{-1}$ serves as a normalization operator for $w_i$, designed to address the scale dependence of sharpness~\cite{DPB17+}, so that it is invariant to any scaling transformation. 
Suppose $S$ is an invertible scaling operator on $\mathbb{R}^k$ that leaves the loss function unchanged, i.e., $\mathcal{L}_i (w_i) = \mathcal{L}_i (S w_i)$. 
On client $i$, for an invertible linear operator $A_{w_i}$ and a given weight vector $w_i$, we have $A_{S w_i}^{-1} S = A_{w_i}^{-1}$. 
Thus, the loss values at $w_i$ and $S w_i$ are identical. 
However, according to the sharpness definition in~\cite{FKM21+}, the sharpness at $w_i$ and $S w_i$ can differ arbitrarily, i.e., 
\begin{equation}
\begin{aligned}
\max_{\parallel \epsilon_i \parallel_2^2 \leq \rho} \mathcal{L}_i &(w_i + \epsilon_i) - \mathcal{L}_i (w_i) \\
&\neq \max_{\parallel \epsilon_i \parallel_2^2 \leq \rho} \mathcal{L}_i (S w_i + \epsilon_i) - \mathcal{L}_i (S w_i), \label{eq:tmp1}
\end{aligned}
\end{equation}
which weakens the correlation between a model's generalization gap and its sharpness. 

To this end, we introduce an adaptive sharpness to train local GNN models as suggested in~\cite{KKP21+}. 
The adaptive sharpness at $w_i$ is defined as: $\max_{\parallel A_{w_i}^{-1} \epsilon_i \parallel_2^2 \leq \rho} \mathcal{L}_i (w_i + \epsilon_i) - \mathcal{L}_i (w_i)$. 
At points $w_i$ and $S w_i$, this adaptive sharpness is identical, i.e., 
\begin{equation}
\begin{aligned}
\max_{\parallel A_{w_i}^{-1} \epsilon_i \parallel_2^2 \leq \rho} \mathcal{L}_i &(w_i + \epsilon_i) - \mathcal{L}_i (w_i) \\
&= \max_{\parallel A_{S w_i}^{-1} \epsilon_i \parallel_2^2 \leq \rho} \mathcal{L}_i (S w_i + \epsilon_i) - \mathcal{L}_i (S w_i). \label{eq:tmp2}
\end{aligned}
\end{equation}

This scale-invariant adaptive sharpness improves the training process in weight space by adjusting maximization regions according to the parameter scale. 
In practice, SEAL solves the min-max problem in Eq.~\eqref{eq:eq5} via a two-step procedure for each update. 
For a small perturbation radius $\rho$, the inner maximization in Eq.~\eqref{eq:eq5} can be approximated by a linearly constrained optimization problem using a first-order Taylor expansion around $w_i$: 
\begin{equation}
\begin{aligned}
A_{w_i^{t}}^{-1} \epsilon_i^{t} &= \mathop{\arg\max}\limits_{ \parallel A_{w_i^{t}}^{-1} \epsilon_i^{t} \parallel_2^2 \leq \rho} \mathcal{L}_i (w_i^{t}  + \epsilon_i^{t}) \\
&\approx \mathop{\arg\max}\limits_{ \parallel A_{w_i^{t}}^{-1} \epsilon_i^{t} \parallel_2^2 \leq \rho} \left( A_{w_i^{t}}^{-1} \epsilon_i^{t} \right)^{\top} A_{w_i^{t}} \nabla \mathcal{L}_i(w_i^{t}) \\
&= \sqrt{\rho} \frac{A_{w_i^{t}} \nabla \mathcal{L}_i(w_i^{t})}{\parallel A_{w_i^{t}} \nabla \mathcal{L}_i(w_i^{t}) \parallel_2}. \label{eq:eq6}
\end{aligned}
\end{equation}

So we have: 
\begin{equation}
\begin{aligned}
\epsilon_i^{t} = \sqrt{\rho} \frac{A_{w_i^{t}}^2 \nabla \mathcal{L}_i(w_i^{t})}{\parallel A_{w_i^{t}} \nabla \mathcal{L}_i(w_i^{t}) \parallel_2}. \label{eq:eq7}
\end{aligned}
\end{equation}

In our implementation, to stabilize the training process, we use $A_{w_i} + \gamma I_{w_i}$ instead of $A_{w_i}$, where $I_{w_i}$ is a diagonal matrix of the same size as $w_i$ with all diagonal entries equal to $1$, and $A_{w_i} = w_i$ is a copy of the current weight matrix detached from the computational graph. 
We take the element-wise absolute value of $A_{w_i}$ and then add $\gamma I_{w_i}$; the resulting matrix is used to rescale the gradient of the current parameter. 
In all experiments, we set $\gamma = 0.1$. 

Subsequently, for each client, the optimizer performs local gradient descent at $w_i^{t}$:
\begin{equation}
\begin{aligned}
w_i^{t+1} = w_i^{t} - \eta_{l} \left( \nabla \mathcal{L}_i (w_i^{t} + \epsilon_i^{t})  + \lambda w_i^{t} \right). \label{eq:eq8}
\end{aligned}
\end{equation} 

From Eq.~\eqref{eq:eq7} and Eq.~\eqref{eq:eq8}, we observe that SEAL first identifies the perturbed point $w_i^{t} + \epsilon_i^{t}$ with the largest local loss within a fixed-radius neighborhood around $w_i^{t}$, and then performs gradient descent at $w_i^{t}$ using the gradient evaluated at $w_i^{t} + \epsilon_i^{t}$.
When $A_{w_i}$ is the identity matrix, the adaptive sharpness reduces to the standard sharpness defined in~\cite{FKM21+}. 
In this case, SEAL no longer enjoys scale invariance; we refer to this variant as \textbf{SEAL-B}, i.e., $\epsilon_i^{t} = \sqrt{\rho} \frac{\nabla \mathcal{L}_i(w_i^{t})}{\parallel \nabla \mathcal{L}_i(w_i^{t}) \parallel_2}$. 

\subsection{Mitigating Dimensional Collapse via Representation Decorrelation}
The primary architectural difference among GNN variants lies in the choice of the aggregation function $\mathrm{Agg} (\cdot)$ in Eq.~\eqref{eq:eq1}, as in GCN~\cite{KW17}, GraphSAGE~\cite{HYL17}, GAT~\cite{VCC18+}, etc. 
These models can be abstracted as
\begin{equation}
    h_v^{l+1} = \sigma \left( W_l \sum_{u \in \mathcal{N}_v}\frac{h_u^{l}}{|\mathcal{N}_v|} + B_l h_v^{l} \right), \forall l \in \left[ L \right], \label{eq:eq9}
\end{equation}
where $\sum_{u \in \mathcal{N}_v} h_u^{l} / |\mathcal{N}_v|$ is the mean of the previous-layer embeddings of the neighbors of node $v$, $W_l$ and $B_l$ are trainable weight and bias matrices, respectively, and $B_l$ can be omitted. 

To analyze the dimensional collapse that may arise during the training process, we consider a simplified $L$-layer GNN ($L \geq 1$) without nonlinear activation functions, following DirectCLR~\cite{JVL22+}. 
Let $h_{i, v}^{L}$ denote the final-layer representation of node $v$ produced by the first $L$ layers of the local GNN on client $i$, and let $\Pi_{i} = W_{i}^{L-1} \cdots W_{i}^{1}W_{i}^{0}$ denote the product of the corresponding $L$ weight matrices. 
For such a simplified GNN, Theorem~\ref{theo:theo1} provides the covariance matrix of graph representations on client $i$. 

\begin{theorem}
Consider a simplified $L$-layer GNN $f_i$, where the update rule is $h_{v}^{l+1} =  W^l_i \sum_{u \in \mathcal{N}_{v}} h_u^{l}$. 
The representation of a graph $G_{i,j}$ on client $i$ is given by $\sum_{v\in V_{i,j}} f(G_{i,j})_v = \Pi_i x^L_{i,j}$, where $x^L_{i,j} = \sum_{v\in \mathcal{V}_{i,j}}  \sum_{u_{L-1} \in \mathcal{N}_{v}}\sum_{u_{L-2} \in \mathcal{N}_{u_{L-1}}} \ldots \sum_{u_0\in \mathcal{N}_{u_1}} x_{u_0}$. 
Accordingly, the covariance matrix of graph representations on client $i$ is 
\begin{equation}
    \Sigma_i = \Pi_i \left( \frac{1}{M_i} \sum_{j=1}^{M_i} \left( x^L_{ i,j} - \bar{x}^L_i \right) \left( x^L_{i,j} - \bar{x}^L_i \right)^{\top} \right) \Pi_i^{\top},
\label{eq:eq10}
\end{equation}
where $\bar{x}^L_i =  \frac{1}{M_i} \sum_{j=1}^{M_i} x^L_{i, j}$.
\label{theo:theo1}
\end{theorem}

The gradient flow analysis of $\Pi_{i}$ indicates that data heterogeneity leads to a significant number of singular values of $\Pi_{i}$ being $0$, which makes $\Pi_{i}$ tend toward a low-rank matrix~\cite{SLZ23+}. 
In FGL, severe data heterogeneity often induces extreme imbalance in the number of local training graph samples per class on each client~\footnote{In the inter-domain setting,  label spaces differ across clients; some classes present on one client may be entirely absent on another.}, so that certain classes may have a vanishingly small (or even zero) proportion of samples. 
This further aggravates the rank reduction of $\Pi_{i}$, leading to dimensional collapse in the resulting graph representations. 
As shown in Theorem~\ref{theo:theo1}, once $\Pi_{i}$ becomes low-rank, the covariance matrix $\Sigma_{i}$ also tends toward low-rank, which results in severe dimensional collapse in local GNN models. 

\begin{algorithm}[htbp]
\caption{SEAL: Sharpness-aware Federated Graph Learning.}
\label{algo:algo1}
\begin{algorithmic}[1]
    \REQUIRE Initial backbone parameters of local models $w_{i}^{0}$, perturbation radius $\rho$, local learning rate $\eta_{l}$, number of local epochs $E$, number of communication rounds $T$, regularization coefficient $\alpha$, hidden dimension $d$. 
    \FOR{$t = 0, \cdots, T-1$}
    \STATE Initialize local backbones with the global backbone: $w_{i, 0}^{t} = w^{t}, \forall i \in [N]$. 
    \FOR{each client $i \in [N]$ in parallel}
    \FOR{$e = 0, \cdots, E-1$}
    \STATE Sample a batch of local data $\xi_i$ and calculate the local loss $\mathcal{L}_i(w_{i, e}^{t}; \xi_{i}) + \alpha \parallel C_i \parallel_{F}^2 / {d^2}$. 
    \STATE Calculate $\nabla_{w_{i, e}} \left( \mathcal{L}_i(w_{i, e}^{t}; \xi_{i}) + \alpha \parallel C_i \parallel_{F}^2 / {d^2} \right)$ to obtain the local gradients and perform the gradient perturbation by Eq.~\eqref{eq:eq7}. 
    \STATE Update the local model according to Eq.~\eqref{eq:eq8}. 
    \ENDFOR
    \ENDFOR
    \STATE Aggregate local backbones on the server: $w^{t+1} = \sum_{i=1}^N \frac{M_i}{M} w_{i}^{t}$. 
    \ENDFOR
\end{algorithmic}
\end{algorithm}

To alleviate this issue, we introduce a regularization term that decorrelates graph representations during local training process. 
Directly regularizing the covariance matrix $\Sigma_i$ would require calculating all of its singular values, which is computationally expensive. 
Instead, we first perform $z$-score normalization on all graph representations $h_{i,j}$: 
\begin{equation}
    \hat{h}_{i, j} = (h_{i, j} - \bar{h}_i) / \sqrt{\text{Var}(h_i)}. \label{eq:eq11}
\end{equation}

We then compute the covariance matrix of the normalized representations, denoted as $C_i$. 
Finally, we add the following regularization term to the local objective: 
\begin{equation}
    \min_{w_i} \mathcal{L}_i(w_i, \xi_i) + \frac{\alpha}{d^2} \parallel C_i \parallel_{F}^2, \label{eq:eq12}
\end{equation}
where $\alpha > 0$ is a regularization coefficient, $d$ is the representation dimension, and $\lVert \cdot \rVert_{F}$ denotes the Frobenius norm, which is computationally efficient. 
The overall training procedure of SEAL is summarized in Algorithm~\ref{algo:algo1}. 

\section{Experimental Results}
\label{sec5}

\subsection{Baseline Setup}

The local-train mode is employed as a baseline to assess whether FGL improves performance through collaborative training. 
In this mode, all clients are initialized with the same GNN model but are trained independently without any communication.
FedAvg~\cite{MMR17+}, FedProx~\cite{LSZ20+}, FedNova~\cite{WLL20+}, and SCAFFOLD~\cite{KKM20+} are selected as federated learning baselines due to their widespread use.
GCFL+~\cite{XMX21+} and FedStar~\cite{TLL23+} are included as state-of-the-art methods specifically designed to handle graph data heterogeneity.
Finally, we also include the non-adaptive sharpness variant of SEAL, which we refer to as SEAL-B.

The experiments are conducted on $16$ graph classification datasets from four domains: \textit{Small Molecules} (AIDS, BZR, COX$2$, DHFR, MUTAG, NCI-$1$, PTC-MR), \textit{Bioinformatics} (DD, ENZYMES, PROTEINS), \textit{Computer Vision} (Letter-high, Letter-med, Letter-low), and \textit{Social Networks} (COLLAB, IMDB-BINARY, IMDB-MULTI).
Node features are available for a subset of these datasets, while graph labels are either binary or multi-class.
We design four FGL settings with increasing levels of data heterogeneity:
\begin{enumerate}
	\item IID Setting: We utilize four benchmark datasets: NCI-$1$, PROTEINS, COLLAB, and IMDB-BINARY.
	In this configuration, data is partitioned such that all clients maintain an identical and balanced label distribution.
	\item Non-IID Setting: Using the same four datasets, we partition the data across clients to introduce label heterogeneity, ensuring that different clients exhibit distinct label distributions.
	\item Cross-Dataset Setting: Each client is randomly assigned a dataset from a collection of seven small-molecule graph datasets.
	We refer to this chemistry-centric configuration as \textit{Chem}.
	\item Inter-Domain Setting: This setting evaluates cross-domain generalization and is further categorized into three variants: \textit{BioChem} (10 datasets from \textit{Small Molecules} and \textit{Bioinformatics}), \textit{BioChemSn} (13 datasets from \textit{Small Molecules}, \textit{Bioinformatics}, and \textit{Social Networks}), and \textit{BioSnCv} (9 datasets from \textit{Bioinformatics}, \textit{Social Networks}, and \textit{Computer Vision}).	
	Here, each client is allocated an independent dataset originating from a distinct domain.
\end{enumerate}
For all configurations, the data is randomly partitioned into training, validation, and test sets with a ratio of $80\%$, $10\%$, and $10\%$, respectively.

All experiments are conducted using a three-layer GAT as the backbone, with results reported as the average over five independent runs using distinct random seeds.
For both IID and non-IID settings, we employ $10$ clients, setting both the hidden dimension and batch size to $64$.
In the more complex \textit{Chem}, \textit{BioChem}, \textit{BioChemSn}, and \textit{BioSnCv} settings, both parameters are adjusted to $32$.
To simulate label heterogeneity in the non-IID setting, we partition the data via a Dirichlet distribution $\mathrm{Dir}_{K}(\beta)$, where $K$ is the number of classes and $\beta = 0.01$ is the concentration parameter. 
Each algorithm is trained for $200$ communication rounds.
Within each round, every client executes one local epoch using an SGD optimizer with a learning rate of $0.003$, weight decay of $10^{-4}$, and momentum of $0.99$.
Hyperparameters are optimized based on validation performance: the regularization coefficient $\alpha$ is determined via grid search over $\{0.001, 0.005, 0.01, 0.1\}$, and the perturbation radius $\rho$ is selected from $\{0.0005, 0.001, 0.005, 0.01\}$.
All experiments are implemented on a single NVIDIA GeForce GTX TITAN $24$GB GPU.

\subsection{Performance Evaluation}

The performance metrics, including average test accuracy, average gain, and the ratio of improved clients, are summarized in Tables~\ref{tab:tab1}--\ref{tab:tab3}.
Specifically, Tables~\ref{tab:tab1} and \ref{tab:tab2} compare all methods under IID and non-IID settings, respectively, while Table~\ref{tab:tab3} details the results for cross-dataset and inter-domain tasks.
A key observation is that algorithms consistently achieve higher accuracy under IID settings compared to non-IID ones (e.g., FedAvg on NCI-$1$).
This indicates that performance is adversely affected by label distribution skew, an effect further validated by the challenging cross-dataset (\textit{Chem}) and inter-domain (\textit{BioChem}) results.

\begin{table*}[htbp]
  \caption{Experimental results under the IID setting. We run experiments on four single datasets and list the results. The average test accuracy ($\%$) over ten clients and the ratio of clients improved are reported. All results are averaged over five runs with different random seeds (mean$\pm$std). The bold font highlights the best performance, while the underlined values indicate the second-best results.}
  \label{tab:tab1}
  \scriptsize
  \centering
  \setstretch{0.75}
  \begin{tabular}{l|ccc|ccc|ccc|ccc}
    \toprule
    \multirow{2}*{Method} & \multicolumn{3}{c}{NCI-$1$} & \multicolumn{3}{c}{PROTEINS} & \multicolumn{3}{c}{COLLAB} & \multicolumn{3}{c}{IMDB-BINARY} \\
    \cmidrule(lr){2-4} \cmidrule(lr){5-7} \cmidrule(lr){8-10} \cmidrule(lr){11-13}
    ~ & test acc & avg. gain & ratio & test acc & avg. gain & ratio & test acc & avg. gain & ratio & test acc & avg. gain & ratio \\
    \midrule
    local-train & 63.97$\pm$0.76 & ~ & ~ & 69.44$\pm$0.53 & ~ & ~ & 70.67$\pm$0.49 & ~ & ~ & 64.67$\pm$0.28 & ~ & ~ \\
    \midrule
    FedAvg & 64.03$\pm$0.41 & 0.06 & 6/10 & 70.01$\pm$0.63 & 0.57 & 6/10 & 70.82$\pm$0.36 & 0.15 & 5/10 & 65.02$\pm$0.67 & 0.35 & 4/10 \\
    +RepDec & 66.27$\pm$0.49 & 2.30 & 6/10 & 72.24$\pm$0.55 & 2.80 & 6/10 & 72.20$\pm$0.32 & 1.53 & 5/10 & 66.31$\pm$0.72 & 1.64 & 5/10 \\
    \midrule
    FedProx & 64.64$\pm$0.54 & 0.67 & 6/10 & 70.83$\pm$0.47 & 1.39 & 7/10 & 71.80$\pm$0.39 & 1.13 & 6/10 & 65.33$\pm$0.46 & 0.66 & 5/10 \\
    +RepDec & 66.19$\pm$0.42 & 2.22 & 6/10 & 73.61$\pm$0.51 & 4.17 & 7/10 & 72.89$\pm$0.46 & 2.22 & 6/10 & 67.34$\pm$0.54 & 2.67 & 5/10 \\   
    \midrule
    SCAFFOLD & 65.71$\pm$0.28 & 1.74 & 4/10 & 70.42$\pm$0.37 & 0.98 & 5/10 & 71.70$\pm$0.41 & 1.03 & 5/10 & 65.36$\pm$0.72 & 0.69 & 5/10 \\
    +RepDec & 66.57$\pm$0.20 & 2.60 & 7/10 & 72.78$\pm$0.33 & 3.34 & 6/10 & 72.47$\pm$0.29 & 1.80 & 5/10 & 66.05$\pm$0.45 & 1.38 & 5/10 \\
    \midrule
    FedNova & 65.12$\pm$0.71 & 1.15 & 4/10 & 70.78$\pm$0.36 & 1.34 & 5/10 & 72.00$\pm$0.67 & 1.33 & 5/10 & 65.25$\pm$0.40 & 0.58 & 5/10 \\
    +RepDec & 66.43$\pm$0.67 & 2.46 & 6/10 & 71.67$\pm$0.29 & 2.23 & 5/10 & 72.50$\pm$0.38 & 1.83 & 6/10 & 65.83$\pm$0.68 & 1.16 & 5/10 \\
    \midrule
    GCFL+ & 65.40$\pm$0.62 & 1.43 & 6/10 & 71.39$\pm$0.59 & 1.95 & 6/10 & 71.80$\pm$0.58 & 1.13 & 6/10 & 65.67$\pm$0.61 & 1.00 & 7/10 \\
    +RepDec & \underline{66.83$\pm$0.54} & \underline{2.86} & 7/10 & 72.50$\pm$0.43 & 3.06 & \underline{8/10} & 72.87$\pm$0.49 & 2.20 & 6/10 & 67.04$\pm$0.81 & 2.37 & 7/10 \\
    \midrule
    FedStar & 65.36$\pm$0.69 & 1.39 & 6/10 & 70.83$\pm$0.23 & 1.39 & 5/10 & 71.05$\pm$0.58 & 0.38 & 6/10 & 65.50$\pm$0.38 & 0.83 & 6/10 \\
    +RepDec & 66.27$\pm$0.62 & 2.30 & \underline{8/10} & 72.51$\pm$0.18 & 3.07 & 7/10 & 72.07$\pm$0.53 & 1.40 & 6/10 & \underline{67.51$\pm$0.36} & \underline{2.84} & 7/10 \\
    \midrule
    SEAL-B w/o RepDec & 65.48$\pm$0.52 & 1.51 & 6/10 & 71.67$\pm$0.52 & 2.23 & 6/10 & 72.13$\pm$0.25 & 1.46 & 6/10 & 66.35$\pm$0.79 & 1.68 & 7/10 \\
    SEAL-B & 66.67$\pm$0.44 & 2.70 & \textbf{8/10} & \textbf{73.89$\pm$0.45} & \textbf{4.45} & 7/10 & \underline{73.40$\pm$0.30} & \underline{2.73} & 6/10 & 67.33$\pm$0.66 & 2.66 & \underline{8/10} \\
    \midrule
    SEAL w/o RepDec & 66.03$\pm$0.42 & 2.06 & 6/10 & 72.26$\pm$0.57 & 2.82 & 7/10 & 72.70$\pm$0.46 & 2.03 & \underline{7/10} & 66.00$\pm$0.38 & 1.33 & 7/10 \\
    SEAL & \textbf{67.62$\pm$0.58} & \textbf{3.65} & 7/10 & \underline{73.33$\pm$0.44} & \underline{3.89} & \textbf{9/10} & \textbf{73.60$\pm$0.55} & \textbf{2.93} & \textbf{8/10} & \textbf{67.69$\pm$0.35} & \textbf{3.02} & \textbf{8/10} \\
  \bottomrule
\end{tabular}
\end{table*}
\begin{table*}[htbp]
  \caption{Experimental results under the non-IID setting.}
  \label{tab:tab2}
  \scriptsize
  \centering
  \setstretch{0.75}
  \begin{tabular}{l|ccc|ccc|ccc|ccc}
    \toprule
    \multirow{2}*{Method} & \multicolumn{3}{c}{NCI-$1$} & \multicolumn{3}{c}{PROTEINS} & \multicolumn{3}{c}{COLLAB} & \multicolumn{3}{c}{IMDB-BINARY} \\
    \cmidrule(lr){2-4} \cmidrule(lr){5-7} \cmidrule(lr){8-10} \cmidrule(lr){11-13}
    ~ & test acc & avg. gain & ratio & test acc & avg. gain & ratio & test acc & avg. gain & ratio & test acc & avg. gain & ratio \\
    \midrule
    local-train & 59.48$\pm$0.65 & ~ & ~ & 62.08$\pm$0.67 & ~ & ~ & 63.21$\pm$0.83 & ~ & ~ & 57.40$\pm$0.43 & ~ & ~\\
    \midrule
    FedAvg & 61.02$\pm$0.52 & 1.54 & 7/10 & 63.13$\pm$0.71 & 1.05 & 5/10 & 64.42$\pm$0.94 & 1.21 & 5/10 & 57.64$\pm$0.32 & 0.24 & 6/10 \\
    +RepDec & 62.21$\pm$0.78 & 2.73 & \textbf{10/10} & 65.34$\pm$0.75 & 3.26 & 5/10 & 66.12$\pm$0.72 & 2.91 & \textbf{7/10} & 58.31$\pm$0.55 & 0.91 & \textbf{9/10} \\
    \midrule
    FedProx & 61.32$\pm$0.56 & 1.84 & 8/10 & 63.31$\pm$0.80 & 1.23 & 5/10 & 65.10$\pm$0.74 & 1.89 & 4/10 & 57.70$\pm$0.66 & 0.30 & 6/10 \\
    +RepDec & 63.86$\pm$0.43 & 4.38 & \textbf{10/10} & 64.42$\pm$0.56 & 2.34 & \underline{6/10} & 66.41$\pm$0.62 & 3.20 & 5/10 & 58.26$\pm$0.54 & 0.86 & \textbf{9/10} \\
    \midrule
    SCAFFOLD & 61.64$\pm$0.27 & 2.16 & 7/10 & 63.37$\pm$0.89 & 1.29 & 5/10 & 65.23$\pm$0.43 & 2.02 & \underline{6/10} & 57.82$\pm$0.61 & 0.42 & 6/10 \\
    +RepDec & 62.19$\pm$0.24 & 2.71 & 7/10 & 64.25$\pm$0.75 & 2.17 & \underline{6/10} & 66.49$\pm$0.21 & 3.28 & \underline{6/10} & 59.42$\pm$0.42 & 2.02 & 7/10 \\
    \midrule
    FedNova & 61.53$\pm$0.55 & 2.05 & 7/10 & 63.58$\pm$0.42 & 1.50 & \underline{6/10} & 65.10$\pm$0.74 & 1.89 & \underline{6/10} & 57.76$\pm$0.57 & 0.36 & 7/10 \\
    +RepDec & 63.25$\pm$0.40 & 3.77 & 7/10 & 64.38$\pm$0.47 & 2.30 & \textbf{7/10} & 66.65$\pm$0.71 & 3.44 & \underline{6/10} & 58.45$\pm$0.53 & 1.05 & 7/10 \\
    \midrule
    GCFL+ & 61.67$\pm$0.84 & 2.19 & \underline{9/10} & 63.81$\pm$0.67 & 1.73 & 4/10 & 65.27$\pm$0.35 & 2.06 & \underline{6/10} & 57.82$\pm$0.52 & 0.42 & 6/10 \\
    +RepDec & \underline{64.95$\pm$0.47} & \underline{5.47} & \textbf{10/10} & 65.81$\pm$0.45 & 3.73 & \underline{6/10} & 67.25$\pm$0.33 & 4.04 & \underline{6/10} & 58.86$\pm$0.41 & 1.46 & 6/10 \\
    \midrule
    FedStar & 62.53$\pm$0.55 & 3.05 & 7/10 & 63.45$\pm$0.55 & 1.37 & 4/10 & 65.78$\pm$0.45 & 2.57 & 5/10 & 58.89$\pm$0.35 & 1.49 & 6/10 \\
    +RepDec & 64.52$\pm$0.56 & 5.04 & \underline{9/10} & 64.64$\pm$0.29 & 2.56 & \textbf{7/10} & 68.83$\pm$0.56 & 5.62 & \underline{6/10} & 60.59$\pm$0.61 & 3.19 & 6/10 \\
    \midrule
    SEAL-B w/o RepDec & 63.03$\pm$0.39 & 3.55 & \underline{9/10} & 63.74$\pm$0.85 & 1.66 & \underline{6/10} & 67.47$\pm$0.66 & 4.26 & \underline{6/10} & 60.72$\pm$0.71 & 3.32 & 6/10 \\
    SEAL-B & 64.86$\pm$0.36 & 5.38 & \underline{9/10} & \underline{66.81$\pm$0.79} & \underline{4.73} & \underline{6/10} & \underline{69.22$\pm$0.40} & \underline{6.01} & \underline{6/10} & \underline{62.17$\pm$0.48} & \underline{4.77} & 6/10 \\
    \midrule
    SEAL w/o RepDec & 64.42$\pm$0.40 & 4.94 & \underline{9/10} & 63.77$\pm$0.55 & 1.69 & \underline{6/10} & 68.87$\pm$0.69 & 5.66 & \underline{6/10} & 60.12$\pm$0.88 & 2.72 & 6/10 \\
    SEAL & \textbf{65.53$\pm$0.70} & \textbf{6.05} & \underline{9/10} & \textbf{67.20$\pm$0.49} & \textbf{5.12} & \underline{6/10} & \textbf{70.69$\pm$0.68} & \textbf{7.48} & \textbf{7/10} & \textbf{63.10$\pm$0.54} & \textbf{5.70} & \underline{8/10} \\
  \bottomrule
\end{tabular}
\end{table*}

Comparing all evaluated methods against the local-train baseline, we observe consistent performance improvements across all experimental settings, as evidenced by the positive average gains (see "avg. gain" columns in Tables~\ref{tab:tab1}--\ref{tab:tab3}).
This confirms that collaborative optimization within the federated framework effectively enhances model performance.
Notably, FedProx, SCAFFOLD, and FedNova exhibit more pronounced gains under non-IID setting compared to IID scenario, suggesting their superior robustness to data heterogeneity relative to the vanilla FedAvg.
Furthermore, in the more challenging cross-dataset and inter-domain settings, GCFL+ and FedStar generally outperform conventional federated learning algorithms, which can be attributed to their specialized architectures tailored for FGL.

As demonstrated in Tables~\ref{tab:tab1} and~\ref{tab:tab2}, the proposed SEAL and SEAL-B consistently outperform all baselines in terms of test accuracy under both IID and non-IID settings.
Table~\ref{tab:tab3} further reveals that SEAL and SEAL-B maintain superior performance in the majority of cases under more challenging cross-dataset and inter-domain settings.
These results validate the efficacy of our approach and highlight its robustness against graph data heterogeneity.
Notably, the integration of proposed Representation Decorrelation (RepDec) consistently improves test accuracy for all evaluated methods, benefiting a significant majority of clients.
This suggests that RepDec effectively enhances local GNN performance by mitigating the dimensional collapse observed in local models (see Fig.~\ref{fig:fig2}), thereby improving the generalization ability of local GNNs across a broad client base.
\begin{table*}[htbp]
  \caption{Experimental results on cross-dataset (Chem) and inter-domain (BioChem, BioChemSn, and BioSnCv) settings.}
  \label{tab:tab3}
  \scriptsize
  \centering
  \setstretch{0.75}
  \begin{tabular}{l|ccc|ccc|ccc|ccc}
    \toprule
    \multirow{2}*{Method} & \multicolumn{3}{c}{Chem} & \multicolumn{3}{c}{BioChem} & \multicolumn{3}{c}{BioChemSn} & \multicolumn{3}{c}{BioSnCv} \\
    \cmidrule(lr){2-4} \cmidrule(lr){5-7} \cmidrule(lr){8-10} \cmidrule(lr){11-13}
    ~ & test acc & avg. gain & ratio & test acc & avg. gain & ratio & test acc & avg. gain & ratio & test acc & avg. gain & ratio \\
    \midrule
    local-train & 72.79$\pm$0.67 & ~ & ~ & 65.71$\pm$0.62 & ~ & ~ & 64.12$\pm$0.45 & ~ & ~ & 65.40$\pm$0.68 & ~ & ~ \\
    \midrule
    FedAvg & 73.34$\pm$0.62 & 0.55 & 4/7 & 66.30$\pm$0.43 & 0.59 & 6/10 & 65.49$\pm$0.53 & 1.37 & 8/13 & 65.68$\pm$0.72 & 0.28 & 5/9 \\
    +RepDec & 75.78$\pm$0.77 & 2.99 & \underline{6/7} & 67.14$\pm$0.69 & 1.43 & 7/10 & 66.36$\pm$0.42 & 2.24 & \underline{10/13} & 66.32$\pm$0.54 & 0.92 & \underline{6/9} \\
    \midrule
    FedProx & 73.42$\pm$0.33 & 0.63 & 4/7 & 67.70$\pm$0.55 & 1.99 & 6/10 & 65.30$\pm$0.52 & 1.18 & 7/13 & 65.88$\pm$0.80 & 0.48 & 5/9 \\
    +RepDec & 76.61$\pm$0.40 & 3.82 & 5/7 & \underline{69.42$\pm$0.58} & \underline{3.71} & \underline{8/10} & 66.24$\pm$0.42 & 2.12 & 8/13 & 66.50$\pm$0.48 & 1.10 & 5/9 \\
    \midrule
    SCAFFOLD & 73.84$\pm$0.52 & 1.05 & 4/7 & 67.63$\pm$0.67 & 1.92 & 5/10 & 65.68$\pm$0.65 & 1.56 & 8/13 & 66.08$\pm$0.71 & 0.68 & \underline{6/9} \\
    +RepDec & 76.64$\pm$0.46 & 3.85 & 5/7 & 67.82$\pm$0.53 & 2.11 & 5/10 & 66.53$\pm$0.54 & 2.41 & 9/13 & 67.14$\pm$0.80 & 1.74 & \textbf{7/9} \\
    \midrule
    FedNova & 73.72$\pm$0.87 & 0.93 & 4/7 & 67.56$\pm$0.81 & 1.85 & 5/10 & 65.50$\pm$0.46 & 1.38 & 8/13 & 66.32$\pm$0.52 & 0.92 & \underline{6/9} \\
    +RepDec & 76.65$\pm$0.83 & 3.86 & 4/7 & 68.35$\pm$0.67 & 2.64 & 5/10 & 66.63$\pm$0.41 & 2.51 & 8/13 & 66.91$\pm$0.56 & 1.51 & \underline{6/9} \\
    \midrule
    GCFL+ & 74.42$\pm$0.68 & 1.63 & 5/7 & 66.88$\pm$0.63 & 1.17 & 6/10 & 65.71$\pm$0.69 & 1.59 & 8/13 & 66.30$\pm$0.58 & 0.90 & 4/9 \\
    +RepDec & 76.43$\pm$0.34 & 3.64 & \underline{6/7} & 68.34$\pm$0.75 & 2.63 & \textbf{9/10} & 67.30$\pm$0.77 & 3.18 & 9/13 & 67.21$\pm$0.53 & 1.81 & \underline{6/9} \\
    \midrule
    FedStar & 75.12$\pm$0.22 & 2.33 & 5/7 & 67.51$\pm$0.69 & 1.80 & 6/10 & 66.21$\pm$0.75 & 2.09 & 7/13 & 65.59$\pm$0.62 & 0.19 & \underline{6/9} \\
    +RepDec & 76.37$\pm$0.74 & 3.58 & \textbf{7/7} & \textbf{69.52$\pm$0.38} & \textbf{3.81} & 6/10 & 67.12$\pm$0.64 & 3.00 & 8/13 & 67.81$\pm$0.37 & 2.41 & \underline{6/9} \\
    \midrule
    SEAL-B w/o RepDec & 75.08$\pm$0.56 & 2.29 & \underline{6/7} & 67.03$\pm$0.46 & 1.32 & 6/10 & 67.54$\pm$0.64 & 3.42 & 8/13 & 67.07$\pm$0.70 & 1.67 & \underline{6/9} \\
    SEAL-B & \underline{76.71$\pm$0.42} & \underline{3.92} & \textbf{7/7} & 68.50$\pm$1.02 & 2.79 & \underline{8/10} & \underline{69.42$\pm$0.74} & \underline{5.30} & 9/13 & \underline{68.01$\pm$0.87} & \underline{2.61} & \underline{6/9} \\
    \midrule
    SEAL w/o RepDec & 75.41$\pm$0.75 & 2.62 & \textbf{7/7} & 67.97$\pm$0.79 & 2.26 & \underline{8/10} & 68.60$\pm$0.44 & 4.48 & 9/13 & 67.84$\pm$0.63 & 2.44 & \underline{6/9} \\
    SEAL & \textbf{77.12$\pm$0.60} & \textbf{4.33} & \textbf{7/7} & 69.37$\pm$0.69 & 3.66 & \underline{8/10} & \textbf{69.71$\pm$0.73} & \textbf{5.59} & \textbf{11/13} & \textbf{68.41$\pm$0.61} & \textbf{3.01} & \textbf{7/9} \\
  \bottomrule
\end{tabular}
\end{table*}
\begin{table*}[htbp]
  \caption{Model update time per round (seconds) and maximum GPU memory usage (GB). We list the single-round model update time and resource consumption of all algorithms during training under different settings.}
  \label{tab:tab4}
  \scriptsize
  \centering
  \setstretch{0.75}
  \begin{tabular}{l|cc|cc|cc|cc|cc}
    \toprule
    \multirow{2}*{Method} & \multicolumn{2}{c}{IID COLLAB} & \multicolumn{2}{c}{Non-IID COLLAB} & \multicolumn{2}{c}{Chem} & \multicolumn{2}{c}{BioChemSn} & \multicolumn{2}{c}{BioSnCv} \\
    \cmidrule(lr){2-3} \cmidrule(lr){4-5} \cmidrule(lr){6-7} \cmidrule(lr){8-9} \cmidrule(lr){10-11}
    ~ & time cost & memory cost & time cost & memory cost & time cost & memory cost & time cost & memory cost & time cost & memory cost \\
    \midrule
    FedAvg & 1.57 & 0.96 & 1.81 & 1.67 & 3.99 & 2.56 & 7.93 & 4.69 & 6.58 & 3.06 \\
    \midrule
    FedProx & 1.67 & 0.96 & 1.89 & 1.67 & 4.06 & 2.59 & 8.07 & 4.75 & 6.72 & 3.24 \\ 
    \midrule
    SCAFFOLD & 2.23 & 1.24 & 2.66 & 1.95 & 4.74 & 2.96 & 9.11 & 4.96 & 6.96 & 3.80 \\
    \midrule
    FedNova & 1.69 & 1.06 & 1.84 & 1.36 & 4.14 & 2.63 & 8.35 & 4.79 & 6.76 & 3.44 \\
    \midrule
    GCFL+ & 1.75 & 1.07 & 1.96 & 1.42 & 4.21 & 2.79 & 8.42 & 4.85 & 6.82 & 3.52 \\
    \midrule
    FedStar & 2.14 & 1.43 & 2.59 & 1.59 & 4.99 & 2.84 & 8.74 & 4.88 & 6.90 & 3.64 \\
    \midrule
    SEAL-B & 2.78 & 1.13 & 3.21 & 1.80 & 7.68 & 2.88 & 17.27 & 4.92 & 8.67 & 3.70 \\
    \midrule
    SEAL & 2.87 & 1.13 & 3.39 & 1.80 & 7.89 & 2.88 & 17.35 & 4.92 & 8.74 & 3.70 \\
  \bottomrule
\end{tabular}
\end{table*}
\begin{figure*}[htbp]
    \subfigure[IID, COLLAB.]{
    \includegraphics[width=0.23\linewidth]{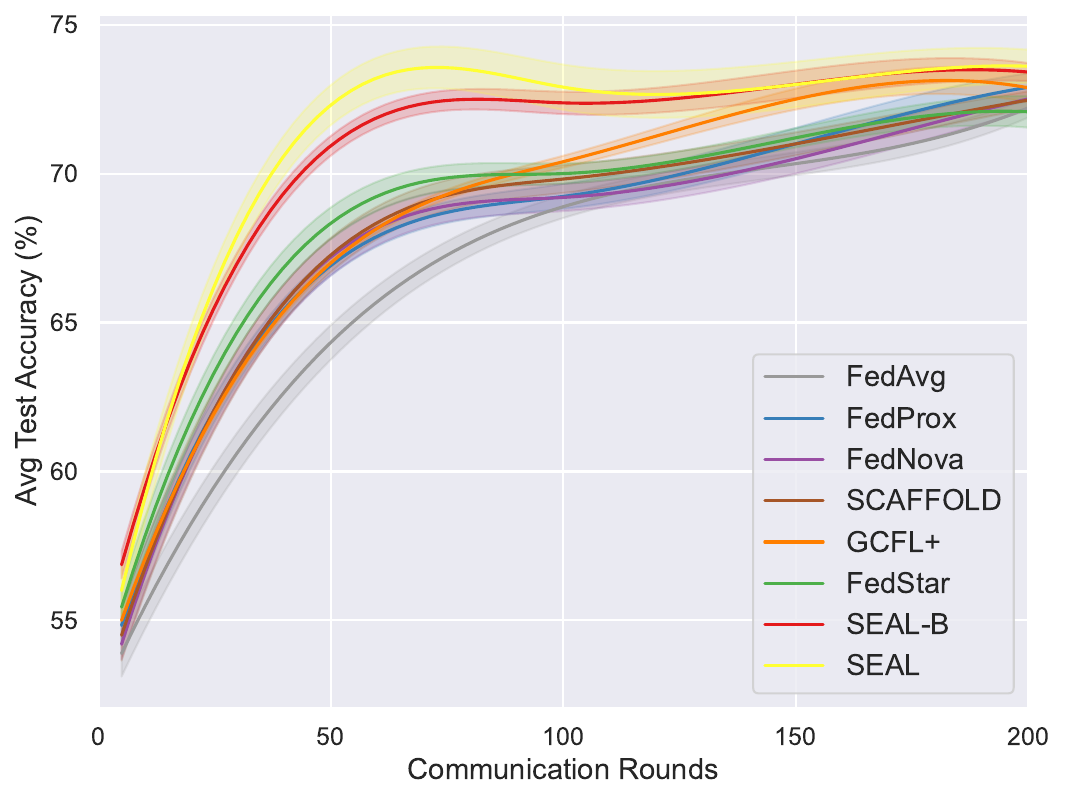}
    \label{fig:fig3a}
    }
    \subfigure[Non-IID, COLLAB.]{
    \includegraphics[width=0.23\linewidth]{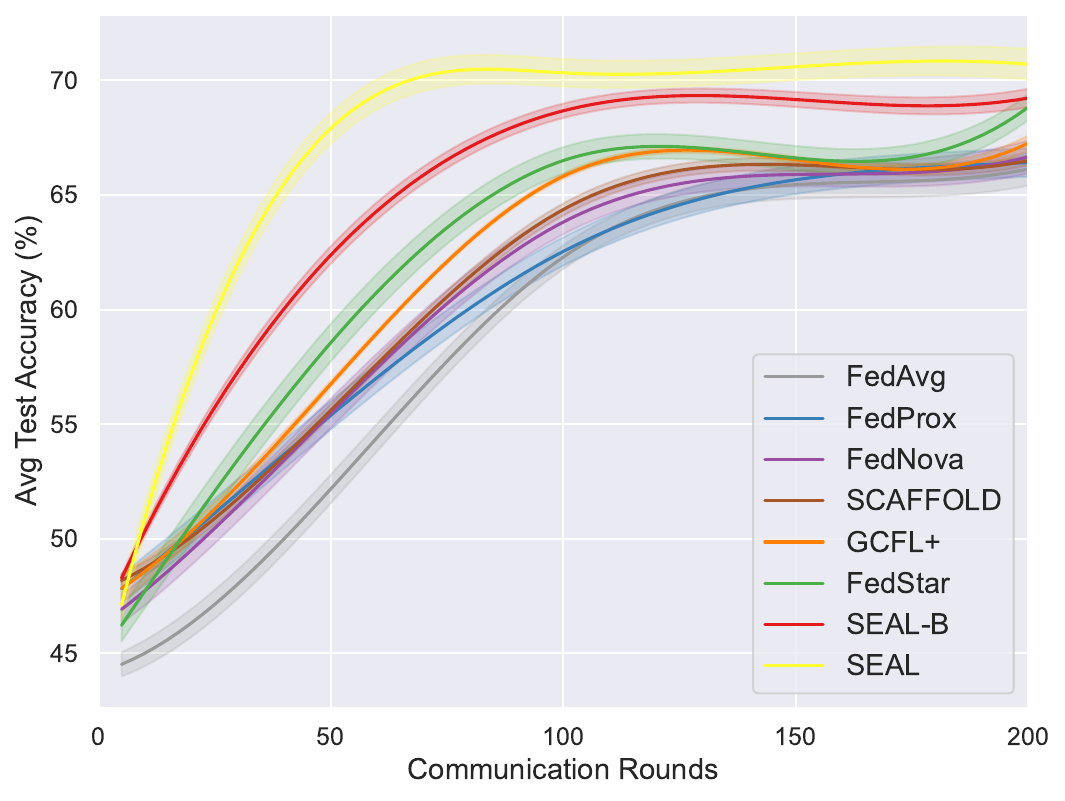}
    \label{fig:fig3b}
    }
    \subfigure[Chem.]{
    \includegraphics[width=0.23\linewidth]{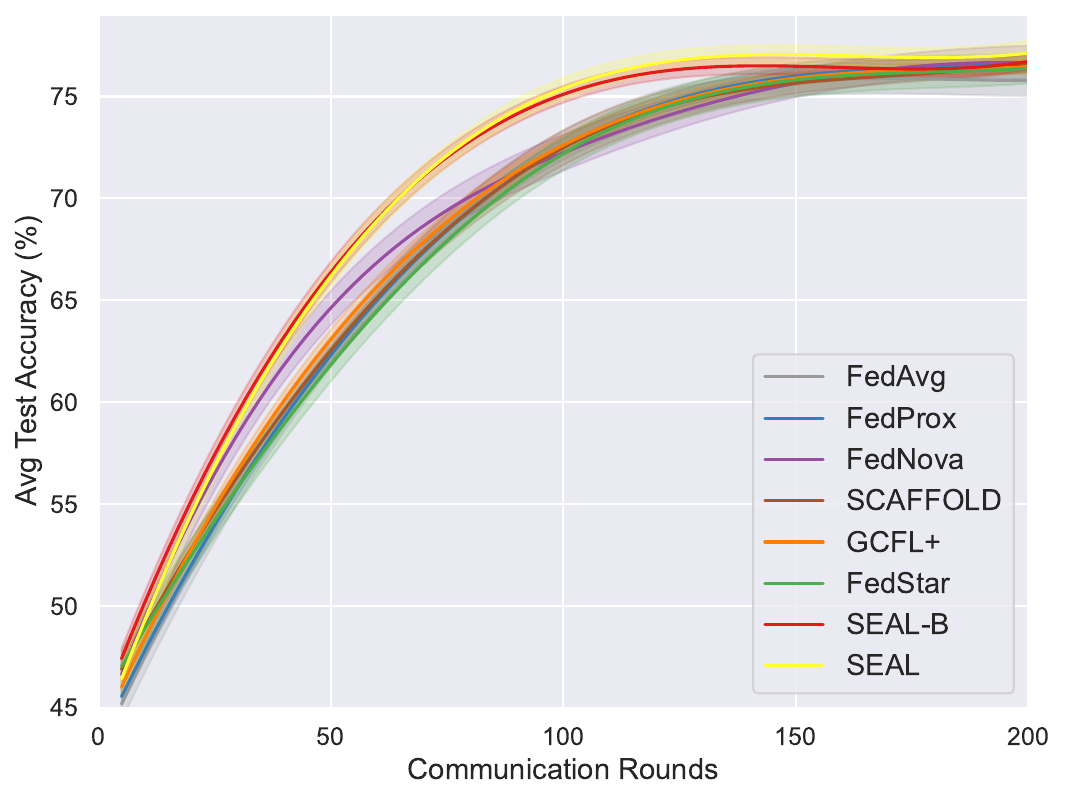}
    \label{fig:fig3c}
    }
    \subfigure[BioSnCV.]{
    \includegraphics[width=0.23\linewidth]{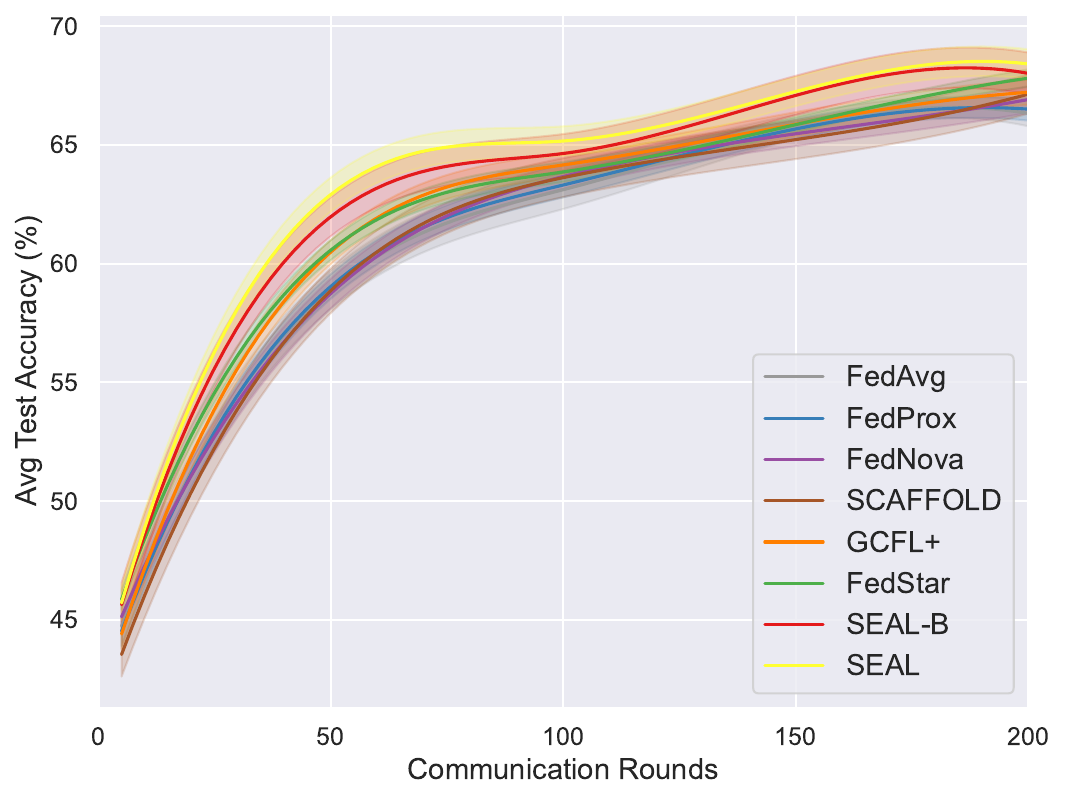}
    \label{fig:fig3d}
    }
    \caption{Test accuracy ($\%$) at each communication round. Results are averaged over five runs. Shaded areas denote one standard deviation above and below the mean values.}
    \label{fig:fig3}
    \centering
\end{figure*}
\begin{figure*}[htbp]
    \subfigure[Non-IID, COLLAB.]{
    \includegraphics[width=0.273\textwidth]{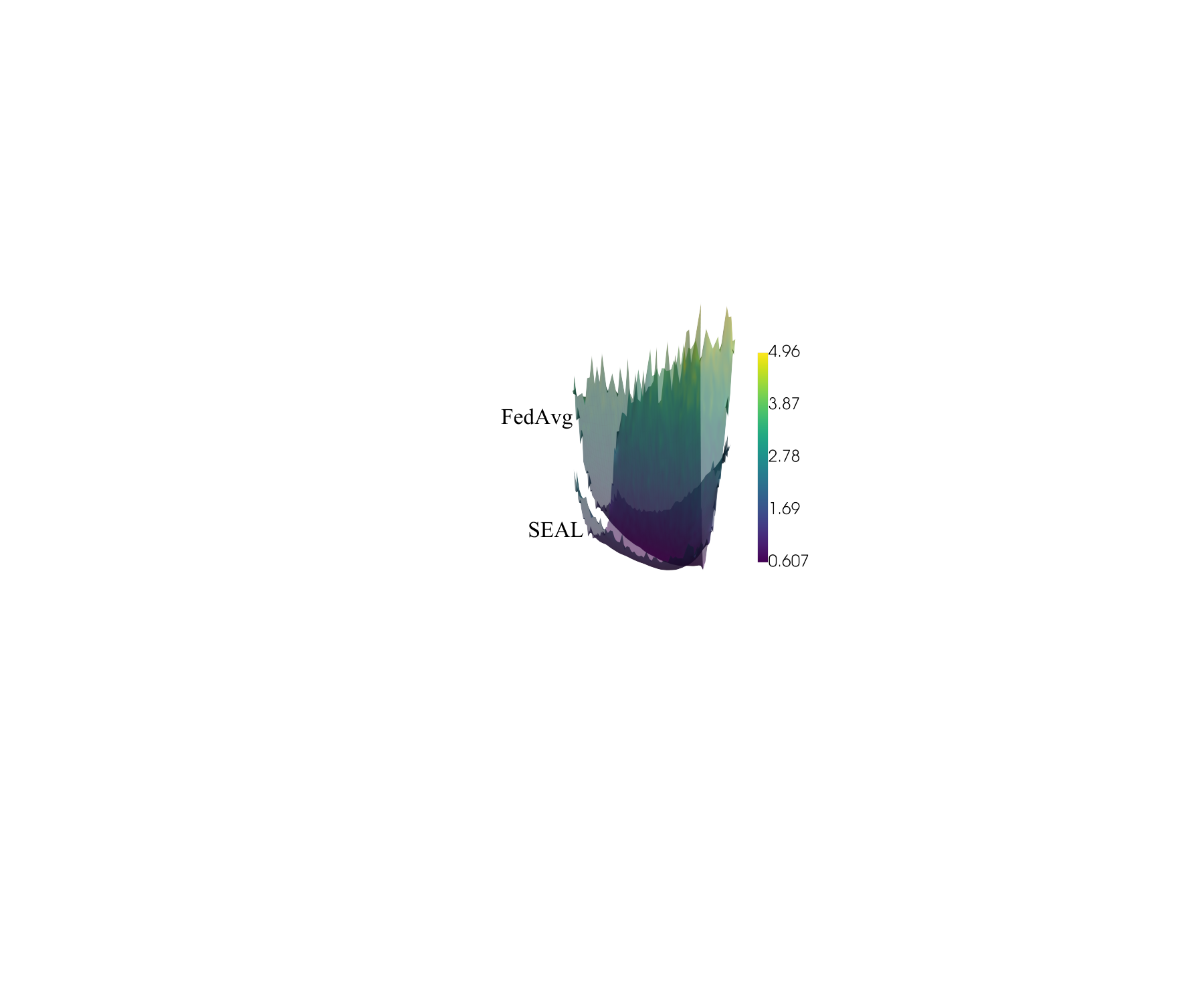}
    \label{fig:fig4a}
    }
    \subfigure[Chem.]{
    \includegraphics[width=0.273\textwidth]{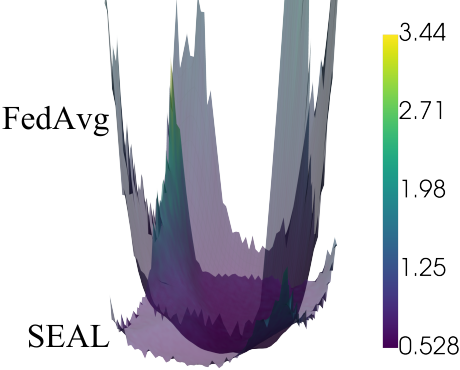}
    \label{fig:fig4b}
    }
    \subfigure[BioChemSn.]{
    \includegraphics[width=0.273\textwidth]{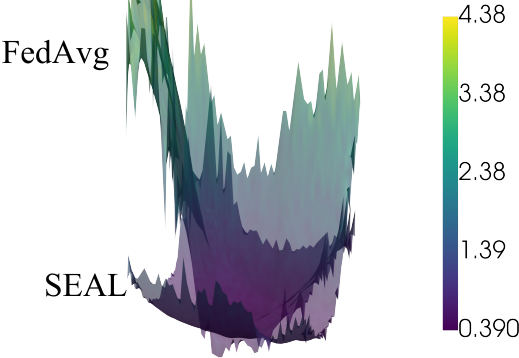}
    \label{fig:fig4c}
    }
    \caption{The Loss landscapes of local GNN models for FedAvg and SEAL under different settings, where models are from the same client that is randomly selected.}
    \label{fig:fig4}
    \centering
\end{figure*}
\begin{figure*}[htbp]
    \subfigure[IID, PROTEINS.]{
    \includegraphics[width=0.23\linewidth]{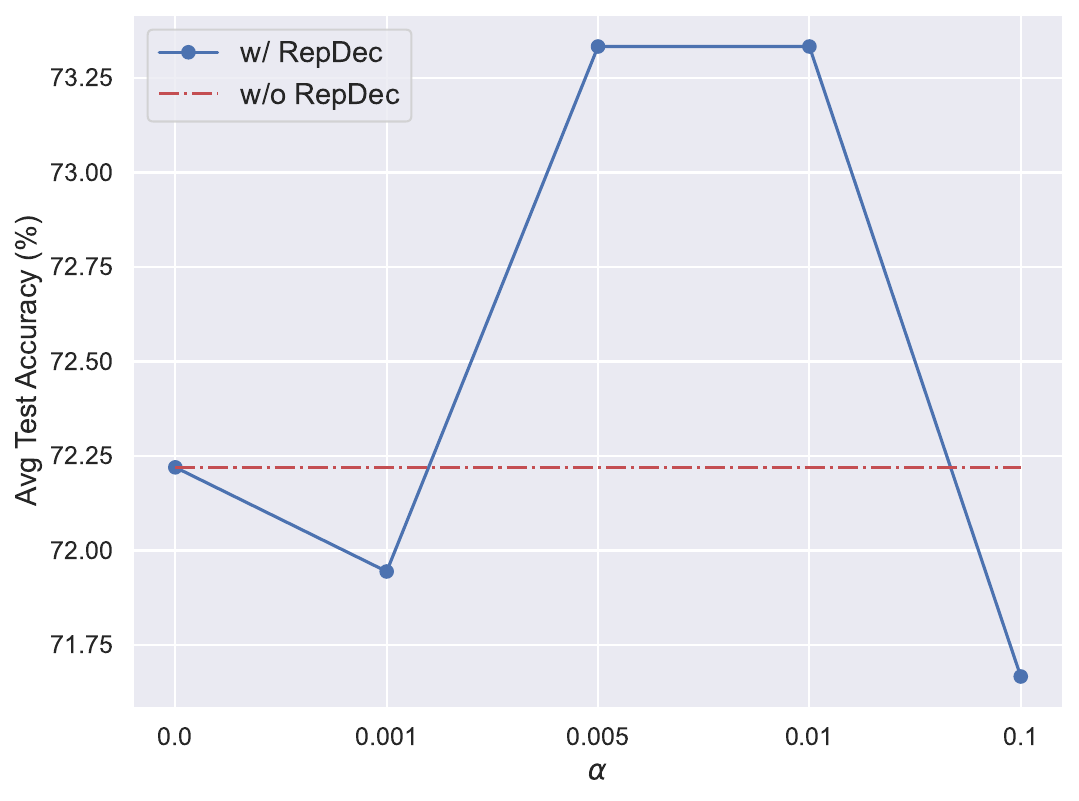}
    \label{fig:fig5a}
    }
    \subfigure[Non-IID, PROTEINS.]{
    \includegraphics[width=0.23\linewidth]{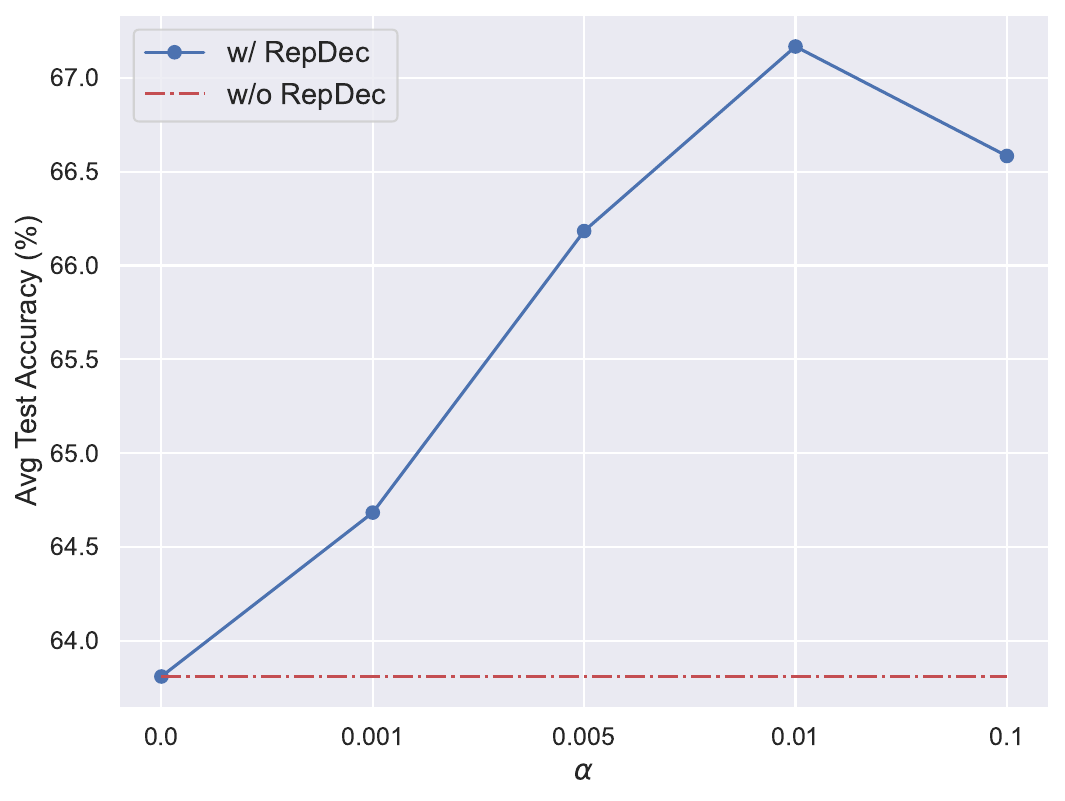}
    \label{fig:fig5b}
    }
    \subfigure[Chem.]{
    \includegraphics[width=0.23\linewidth]{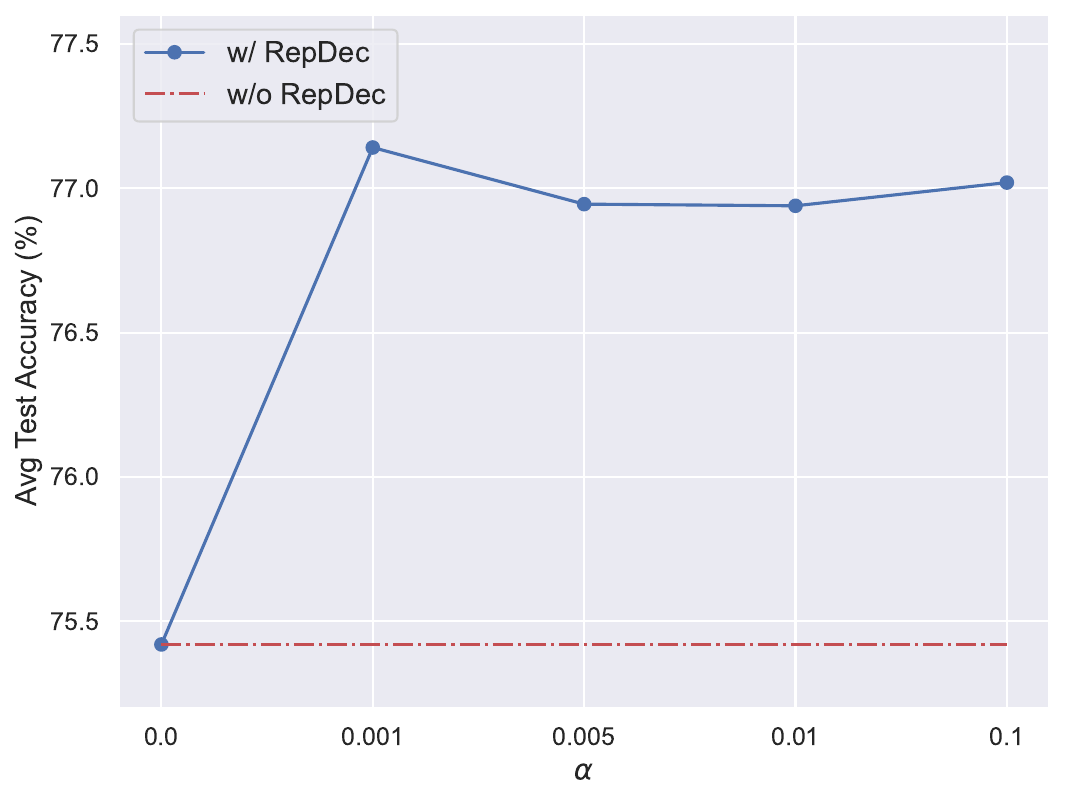}
    \label{fig:fig5c}
    }
    \subfigure[BioChem.]{
    \includegraphics[width=0.23\linewidth]{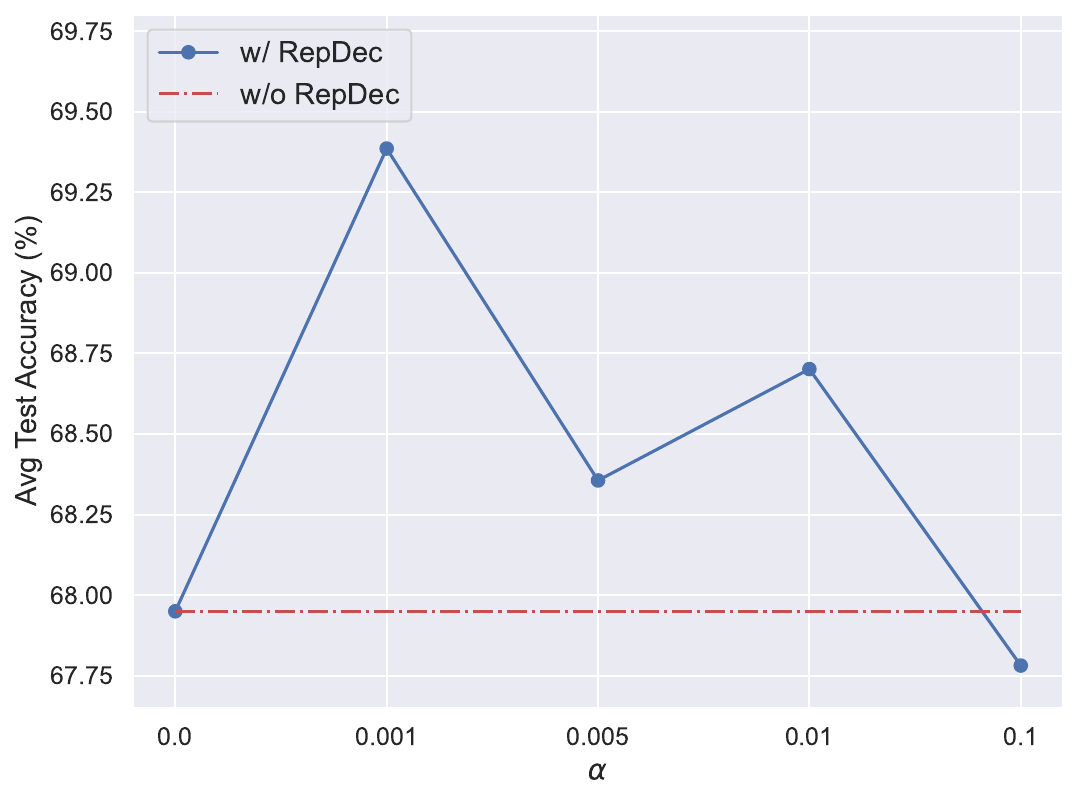}
    \label{fig:fig5d}
    }
    \caption{Impact of regularizer coefficient $\alpha$.}
    \label{fig:fig5}
    \centering
\end{figure*}

To evaluate convergence performance, we plot the evolution of the average test accuracy (including standard deviation over five independent runs) across the entire training process in Fig.~\ref{fig:fig3}.
It is observed that SEAL and SEAL-B consistently outperform the baselines in terms of both final accuracy and convergence rate (i.e., requiring fewer communication rounds to achieve a target performance level) across all configurations.
Furthermore, Fig.~\ref{fig:fig4} visualizes the loss landscapes of local GNNs from a representative client under three distinct settings.
Compared to FedAvg, SEAL yields a significantly flatter loss landscape.
This increased flatness suggests that the models trained with SEAL possess superior generalization capabilities.
It is worth noting that SEAL does not necessarily converge to the absolute global minimum loss.
For instance, in the \textit{BioChemSn} setting, FedAvg attains a lower training loss than SEAL.
This phenomenon stems from the fact that SEAL optimizes a joint objective that balances loss magnitude with sharpness, rather than minimizing loss in isolation.
Finally, Table~\ref{tab:tab4} summarizes the average model update time per round and the peak GPU memory footprint for each algorithm.
Since SAM necessitates two forward-backward passes and additional gradient storage, SEAL and SEAL-B exhibit higher computational overhead and memory consumption compared to the other baselines. 

\subsection{Ablation Analysis}
\subsubsection{Impact of regularizer coefficient $\alpha$.}
We study the impact of $\alpha$ in Eq.~\eqref{eq:eq12} by selecting ${0.001, 0.005, 0.01, 0.1}$, and report the results of SEAL under four different settings in Fig.~\ref{fig:fig5}. 
For the non-IID setting, we partition the PROTEINS dataset across ten clients using $\mathrm{Dir}_{2}(0.01)$. 
As $\alpha$ increases, performance first improves and then degrades, suggesting that $\alpha$ must be selected within a suitable range to achieve strong performance. 
In our experiments, setting $\alpha$ to $0.005$ or $0.01$ yields (near-)optimal performance across settings; therefore, we recommend selecting $\alpha$ in the range $0.005 \leq \alpha \leq 0.01$. 

\subsubsection{Impact of perturbation radius $\rho$.}
To investigate the impact of $\rho$ on SEAL, we vary its value in $\left\{0.0005, 0.001, 0.005, 0.01\right\}$ and summarize the results in Table~\ref{tab:tab5}. 
As before, the PROTEINS dataset is partitioned across ten clients using $\mathrm{Dir}_{2}(0.01)$ to simulate a non-IID label distribution. 
As $\rho$ increases, the average test accuracy initially improves and then saturates (or slightly decreases), indicating that different settings require carefully tuned values of $\rho$ to achieve the best performance. 
In our experiments, we therefore set $\rho = 0.005$ for the IID and Non-IID settings on the four single datasets, and $\rho = 0.001$ for cross-dataset and inter-domain settings.

\begin{table}[htbp]
  \caption{Impact of perturbation radius $\rho$.}
  \label{tab:tab5}
  \footnotesize
  \centering
  \begin{tabular}{lcccc}
    \toprule
    $\rho$ & 0.0005 & 0.001 & 0.005 & 0.01 \\
    \midrule
    IID & 72.51$\pm$0.62 & 72.78$\pm$0.58 & \textbf{73.33$\pm$0.45} & 73.06$\pm$0.65 \\
    Non-IID & 64.58$\pm$0.76 & 66.18$\pm$0.62 & \textbf{67.16$\pm$0.37} & 66.24$\pm$0.49 \\
    Chem & 75.80$\pm$0.78 & \textbf{77. 14$\pm$0.53} & 76.96$\pm$0.62 & 76.08$\pm$0.48 \\
    BioChem & 68.41$\pm$0.67 & \textbf{69.39$\pm$0.71} & 68.89$\pm$0.59 & 69.27$\pm$0.54 \\
  \bottomrule
\end{tabular}
\end{table}

\begin{table}[htbp]
  \caption{Impact of local epochs $E$.}
  \label{tab:tab6}
  \footnotesize
  \centering
  \begin{tabular}{lcccc}
    \toprule
    $E$ & 1 & 2 & 3 & 4 \\
    \midrule
    FedAvg & 75.81$\pm$0.72 & \textbf{78.80$\pm$0.61} & 77.98$\pm$0.60 & 77.75$\pm$0.46 \\
    FedNova & 75.54$\pm$0.38 & 75.42$\pm$0.55 & 75.14$\pm$0.71 & \textbf{76.50$\pm$0.54} \\
    FedStar & 76.38$\pm$0.33 & 76.93$\pm$0.39 & \textbf{78.93$\pm$0.38} & 77.26$\pm$0.21 \\
    SEAL-B & 76.69$\pm$0.61 & \textbf{78.36$\pm$0.55} & 77.97$\pm$0.37 & 77.68$\pm$0.44 \\
    SEAL & 77.14$\pm$0.62 & 78.55$\pm$0.50 & \textbf{79.08$\pm$0.56} & 78.92$\pm$0.47 \\
  \bottomrule
\end{tabular}
\end{table}

\subsubsection{Impact of local epochs $E$.}
We vary the number of local epochs $E$ per communication round in $\left\{ 1, 2, 3, 4 \right\}$ for five different algorithms to investigate its impact. 
For fairness, we fix $\alpha = \rho = 0.001$ throughout. 
The results in Table~\ref{tab:tab6} show that SEAL-B and SEAL outperform the other methods in most cases. 
As $E$ increases, the performance of all five algorithms first improves, then saturates, and finally drops slightly. 
This behavior arises because, when $E$ is too small, local training cannot sufficiently approach the local optimum within each communication round, whereas an overly large $E$ may cause local models to drift too far from the global optimum. 
In all our experiments, we therefore set the default value of $E$ to $1$.

\section{Conclusion}
\label{sec6}
In this work, we propose a novel federated graph learning algorithm, SEAL, specifically designed to effectively address the challenge of graph data heterogeneity. 
We aim to improve the generalization ability of as many local GNN models as possible during training from two different perspectives. 
On the one hand, we incorporate a sharpness-aware minimization optimizer into each local GNN, which jointly minimizes the local loss and its sharpness so as to search for model parameters in a flat region of the loss landscape with uniformly low loss values. 
On the other hand, we decorrelate local graph representations to mitigate the dimensional collapse observed across clients, thereby preserving richer and more discriminative features for downstream classification tasks. 
Extensive experiments on a wide range of graph classification benchmarks demonstrate that SEAL consistently outperforms state-of-the-art federated graph classification methods and exhibits better generalization under diverse and heterogeneous graph data distribution settings. 

\section{Acknowledgments}
\label{sec7}

This work was supported by the major research plan of the National Natural Science Foundation of China (Grant No. 92267204).

\bibliographystyle{ACM-Reference-Format}
\bibliography{sample-base}


\end{document}